%% file: main.tex
\documentclass{article}
\usepackage{iclr2027_conference,times}

\usepackage[utf8]{inputenc}
\usepackage[T1]{fontenc}
\usepackage{amsmath,amssymb}

\usepackage{graphicx}
\usepackage{booktabs}
\usepackage{multirow}
\usepackage[table]{xcolor}
\usepackage{microtype}
\usepackage{xspace}
\usepackage{subcaption}
\usepackage{wrapfig}
\usepackage{enumitem}

\usepackage{algorithm}
\usepackage[noend]{algpseudocode}

\usepackage{url}
\usepackage{hyperref}
\usepackage[capitalize]{cleveref}

\input{math_commands.tex}

\definecolor{rankfirst}{HTML}{F4CCCC}
\definecolor{ranksecond}{HTML}{FCE5CD}
\definecolor{rankthird}{HTML}{FFF2CC}
\definecolor{bestcolor}{RGB}{220,245,220}
\definecolor{secondcolor}{RGB}{255,242,204}

\newcommand{\first}[1]{\cellcolor{rankfirst}\textbf{#1}}
\newcommand{\second}[1]{\cellcolor{ranksecond}#1}
\newcommand{\third}[1]{\cellcolor{rankthird}#1}
\newcommand{\best}[1]{\cellcolor{bestcolor}\textbf{#1}}

\newcommand{\rwhead}[1]{\vspace{0.3em}\noindent\textbf{#1.}\xspace}

\title{Multi-Horizon Future Discovery for Physics-Informed Exploration}

\author{%
\begin{minipage}[t]{\dimexpr\textwidth-2\tabcolsep\relax}%
\centering\normalfont
\textbf{Shuwei Wang}$^{1,2,3}$\hspace{0.8em}\textbf{Qiang Guan}$^{1,2,3,\dagger}$\hspace{0.8em}\textbf{Brian Qiu}$^{1}$\hspace{0.8em}\textbf{Haoxiang Huang}$^{4}$\\[2pt]
\textbf{Haoxi Wang}$^{1,2}$\hspace{0.8em}\textbf{Chenyu Li}$^{1,2}$\hspace{0.8em}\textbf{Bo Xu}$^{1,2,3,\dagger}$\hspace{0.8em}\textbf{Sen Cui}$^{5,\dagger}$\\[6pt]
{\small
$^{1}$University of Chinese Academy of Sciences\\
$^{2}$Institute of Automation, Chinese Academy of Sciences (CASIA)\\
$^{3}$School of Artificial Intelligence, University of Chinese Academy of Sciences\\
$^{4}$University of Science and Technology of China\\
$^{5}$Tsinghua University\\[4pt]
$^{\dagger}$Corresponding authors
}%
\end{minipage}%
}

\iclrfinalcopy

\begin{document}
\maketitle

\fancyhead{}
\renewcommand{\headrulewidth}{0pt}

% ============================================================
% Abstract
% ============================================================
\begin{abstract}
Exploration in physical control environments requires agents to identify actions whose consequences reveal previously unexplored aspects of the underlying dynamics. However, many intrinsic motivation methods rely on immediate prediction errors, state novelty, or local physical changes, which may overlook informative outcomes that emerge across extended temporal horizons. We investigate whether learning structured representations of reachable futures can provide a more effective basis for exploration. To this end, we introduce Hamiltonian Temporal Contrastive Curiosity (HTC$^2$), a framework that combines multi-horizon temporal representation learning with Hamiltonian phase-space structure. A horizon-conditioned contrastive critic learns temporal relationships from future-state pairs sampled within the same episode. The distance between positive-pair representations is then used as an intrinsic reward, encouraging the agent to explore futures that remain insufficiently organized in the learned representation. To incorporate information about physical dynamics, we additionally modulate this reward using bounded multi-step Hamiltonian variation, rather than directly rewarding changes in the Hamiltonian itself. We integrate HTC$^2$ with both Soft Actor-Critic (SAC) and Proximal Policy Optimization (PPO), and evaluate it across six DeepMind Control Suite tasks. Comparisons with ICM, RIDE, E3B, Plan2Explore, and a single-step Hamiltonian curiosity baseline are complemented by state-coverage evaluations, component ablations, and phase-space diagnostics to characterize the contributions of temporal learning and physical structure to exploratory behavior.
\end{abstract}

% ============================================================
% Main text
% ============================================================
\input{main/1_introduction}
\input{main/2_related_work}
\input{main/3_method}
\input{main/4_experiments}
\input{main/5_conclusion}

% ============================================================
% AI use statement
% ============================================================
\section*{AI Use Statement}
During the preparation of this manuscript, the authors used ChatGPT to assist with language polishing, including grammar correction and phrasing improvements. ChatGPT was not used for research ideation. The authors reviewed all AI-assisted text and take full responsibility for the final content of this work.

% ============================================================
% References
% ============================================================
{\small
\bibliographystyle{iclr2027_conference}
\bibliography{main}
}

% ============================================================
% Appendix
% ============================================================
\newpage
\input{main/6_appendix}

\end{document}

%% file: math_commands.tex
\usepackage{amsmath,amsfonts,bm}

\def\eqref#1{equation~\ref{#1}}
\def\1{\bm{1}}

\DeclareMathAlphabet{\mathsfit}{\encodingdefault}{\sfdefault}{m}{sl}
\SetMathAlphabet{\mathsfit}{bold}{\encodingdefault}{\sfdefault}{bx}{n}

%% file: main/1_introduction.tex
\section{Introduction}
\label{sec:intro}

\begin{figure}[t]
    \centering
    \includegraphics[width=\linewidth]{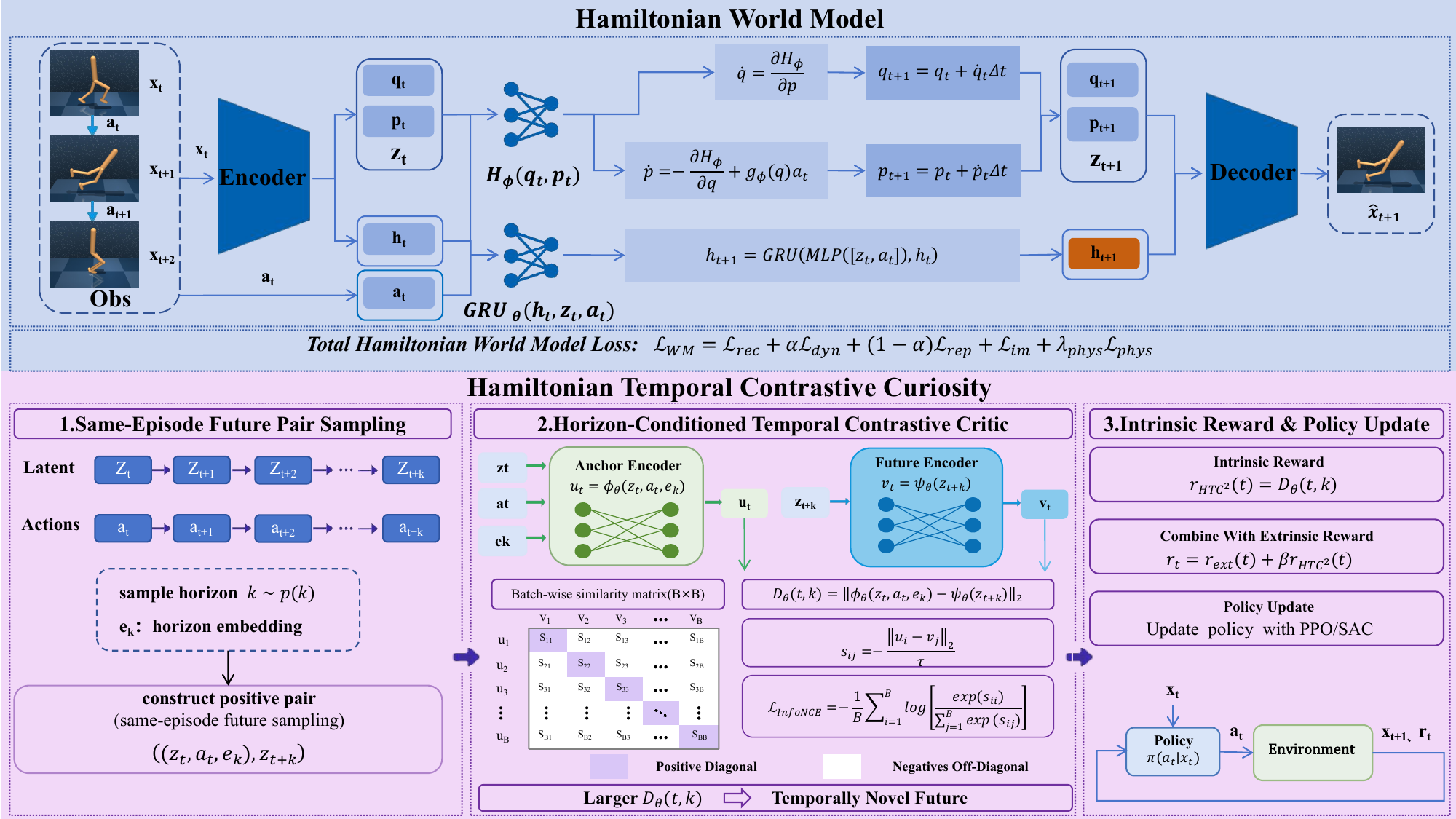}
    \caption{Overview of \textbf{Hamiltonian Temporal Contrastive Curiosity} (HTC$^2$). Observations are encoded into Hamiltonian phase states; same-episode future pairs train a horizon-conditioned temporal contrastive critic; and the resulting temporal discovery signal is modulated by bounded multi-step Hamiltonian variation.}
    \label{fig:overview}
\end{figure}

Exploration is a central challenge in reinforcement learning (RL), especially in physical control, where useful behaviors often emerge only through temporally extended interactions with the environment~\citep{burda2018exploration,pathak2017curiosity,thrun1992efficient,stadie2015incentivizing,sutton1998reinforcement,mohamed2026temporal}. In such settings, an agent must discover not only unfamiliar states, but also how its actions shape future system dynamics~\citep{toth2019hamiltonian,garcia2025learning}. For example, locomotion depends on coordinated changes in balance, contact, velocity, and momentum over multiple time steps rather than merely visiting visually different postures~\citep{arzani2025interpreting,noda2025advancing,toth2019hamiltonian,garcia2025learning}. Physical exploration therefore requires more than generic state-space coverage~\citep{pathak2017curiosity,mohamed2026temporal,burda2018exploration,raileanu2020ride}: the agent must identify action-conditioned futures that reveal informative dynamics.

Existing curiosity-driven methods promote exploration through prediction error, novelty, episodic uncertainty, or representation displacement~\citep{pathak2017curiosity,burda2018exploration,raileanu2020ride,henaff2022exploration}. These signals are often defined from local states or short transitions: prediction error can be affected by irrelevant stochasticity, novelty may encourage visitation without distinguishing dynamical relevance, and representation displacement does not necessarily separate coherent physical evolution from arbitrary feature change. Temporal contrastive learning provides a way to capture longer-range relations by learning representations between current state--action pairs and their future states~\citep{temporal_contrastive_exploration,hansen2022temporal}. Such relations can identify futures that remain poorly captured across multiple temporal scales, but they do not by themselves encode structured variation in the underlying physical dynamics.

Physical inductive biases provide a natural way to structure this temporal information. Hamiltonian world models represent dynamics through coordinate-like and momentum-like latent variables and organize their evolution with a learned Hamiltonian~\citep{greydanus2019hamiltonian,toth2019hamiltonian,tang2026dreamsac}. Such phase-space representations describe how actions alter latent dynamics over time. DreamSAC exploits this structure by rewarding the change in the learned Hamiltonian across a single transition~\citep{tang2026dreamsac}. However, single-step Hamiltonian-change curiosity remains local in time: a large instantaneous variation does not necessarily indicate that an action reaches an informative future over longer horizons. This motivates exploration that identifies informative action-conditioned futures across multiple temporal scales while accounting for structured variation in the learned dynamics.

To address this challenge, we introduce \textbf{Hamiltonian Temporal Contrastive Curiosity} (\textbf{HTC}$^2$), which formulates exploration as multi-step future discovery in a learned Hamiltonian phase space. As illustrated in Figure~\ref{fig:overview}, HTC$^2$ trains a horizon-conditioned temporal contrastive critic on same-episode future pairs, using positive-pair representation distance as the primary curiosity signal. Bounded multi-step Hamiltonian variation then modulates this signal according to structured changes in the learned phase dynamics, while temporal novelty remains the exploration objective. A latent RND~\citep{burda2018exploration} warm start stabilizes the initial learning stage before transitioning to the HTC$^2$ signal.

Our contributions are summarized as follows:
\begin{enumerate}[leftmargin=2.5em,itemsep=2pt,topsep=2pt,parsep=0pt,partopsep=0pt]
    \item We formulate physical exploration as multi-step action-conditioned future discovery in a learned Hamiltonian phase space, moving beyond curiosity based only on local novelty or single-step Hamiltonian change.
    \item We introduce HTC$^2$, using horizon-conditioned temporal contrastive distance as the primary curiosity signal and bounded multi-step Hamiltonian variation as structural modulation of learned phase dynamics.
    \item We provide comprehensive empirical validation on DeepMind Control Suite tasks with SAC and PPO against ICM, RIDE, E3B, Plan2Explore, and a single-step Hamiltonian-change variant, covering control performance, physical-state coverage, parametric OOD robustness, ablations, and Hamiltonian diagnostics.
\end{enumerate}

%% file: main/2_related_work.tex
\section{Related Work}
\label{sec:related}

\rwhead{Exploration in Reinforcement Learning}
Exploration remains a central challenge in reinforcement learning, especially under sparse or delayed rewards ~\citep{sutton1998reinforcement,thrun1992efficient,mohamed2026temporal}. Intrinsic-motivation methods address this problem by rewarding novelty, uncertainty, or prediction difficulty ~\citep{houthooft2016vime,ostrovski2017count,burda2018exploration,pathak2017curiosity,burda2018large,lee2019efficient,guo2022byol}. Count-based approaches favor rarely visited states ~\citep{bellemare2016unifying,tang2017exploration,ostrovski2017count}, but scale poorly to high-dimensional continuous domains. Prediction-based methods instead use learned errors or uncertainty: ICM relies on forward prediction error~\citep{pathak2017curiosity}, RND measures error against a fixed random target~\citep{burda2018exploration}, and related methods exploit representation change, model disagreement, information gain, or episodic uncertainty ~\citep{raileanu2020ride,pathak2019self,houthooft2016vime,henaff2022exploration}. Such local novelty signals can be affected by irrelevant stochasticity, as illustrated by the ``noisy-TV'' problem ~\citep{pathak2019self,mavor2022stay}. More recently, model-based exploration has leveraged learned world models to plan toward informative future states and improve sample efficiency~\citep{sekar2020planning}. However, such models are primarily learned as predictive or generative dynamics models and do not explicitly impose Hamiltonian structure or related physics-informed priors on the latent dynamics.

\rwhead{World Models in Reinforcement Learning}
World models learn predictive representations of environment dynamics and are widely used in model-based reinforcement learning ~\citep{ha2018world,hafner2020mastering,lyu2025dywa,hafner2023mastering,hafner2025mastering,xu2022learning,liu2023learning,ha2018recurrent,hafner2019learning,zhang2026world}. Latent world models compress high-dimensional observations and predict future latent states for planning, imagination, or policy learning ~\citep{hafner2019dream,assran2025v,assran2023self}. The Dreamer family uses recurrent latent dynamics with predictive reconstruction objectives ~\citep{hafner2020mastering,hafner2023mastering,hafner2025mastering}, whereas TD-MPC-style methods emphasize task-relevant latent prediction for efficient control ~\citep{hansen2024td,hansen2022temporal,janner2019trust}. Learned dynamics are also used for intrinsic exploration through prediction error, inverse dynamics, and uncertainty ~\citep{pathak2017curiosity,pathak2019self,raileanu2020ride,henaff2022exploration}. However, standard world models mainly capture statistical regularities and need not explicitly encode physical structure ~\citep{hafner2020mastering,hafner2023mastering,hafner2025mastering,hafner2019dream,assran2025v,assran2023self}, nor does predictive modeling alone specify which future interactions are useful for exploration~\citep{sekar2020planning,tang2026dreamsac}. HTC$^2$ uses a Hamiltonian phase representation as the space for measuring temporal future relations, turning the world model into a structured substrate for intrinsic exploration.

\rwhead{Hamiltonian Dynamics for World Models}
Physics-structured world models introduce inductive biases from classical mechanics to improve dynamics learning and generalization ~\citep{greydanus2019hamiltonian,cranmer2020lagrangian,zhong2019symplectic,toth2019hamiltonian,arzani2025interpreting,troch2025action}. Hamiltonian neural networks and symplectic models represent dynamics through coordinate-like and momentum-like variables governed by gradients of a learned Hamiltonian ~\citep{toth2019hamiltonian,greydanus2019hamiltonian,zhong2019symplectic}. Recent work extends these priors to reinforcement learning and pixel-based world models ~\citep{schiewer2024exploring,deng2025denoising,tang2026dreamsac}. Their use for exploration, however, remains less developed; directly rewarding learned energy change can favor local perturbations rather than temporally extended discovery ~\citep{tang2026dreamsac}. HTC$^2$ instead uses Hamiltonian variation only as a bounded modulation, while the primary exploration signal is the temporal contrastive distance between current phase state--action pairs and their multi-step futures.

%% file: main/3_method.tex
\section{Method}
\label{sec:method}

We propose \textbf{Hamiltonian Temporal Contrastive Curiosity} (\textbf{HTC}$^2$), an intrinsic exploration method for reinforcement learning with Hamiltonian latent world models. HTC$^2$ encourages the agent to discover future phase states that remain difficult to organize under the current temporal representation, while using learned Hamiltonian variation to provide a bounded physics-guided modulation. As illustrated in Figure~\ref{fig:overview}, HTC$^2$ consists of a Hamiltonian phase-space world model, a horizon-conditioned temporal contrastive critic, and a temporal discovery reward with latent RND warm start.

\subsection{Preliminaries}
\label{sec:preliminaries}

\rwhead{Controlled Hamiltonian Dynamics}
We consider latent dynamics with a phase state $z_t=(q_t,p_t)$, where $q_t$ and $p_t$ are learned coordinate-like and momentum-like latent variables~\citep{toth2019hamiltonian}. A controlled Hamiltonian formulation separates the internal latent dynamics from the effect of external actions~\citep{zhong2019symplectic}. Given a learned Hamiltonian $H_\varphi(q,p)$ and a learned control input mapping $g_\varphi(q)$, the continuous-time dynamics are

\begin{equation}
\label{eq:controlled_hamiltonian}
    \frac{dq}{dt}=\frac{\partial H_\varphi(q,p)}{\partial p},
    \qquad
    \frac{dp}{dt}=-\frac{\partial H_\varphi(q,p)}{\partial q}
    +g_\varphi(q)a_t .
\end{equation}
The first component describes the internal Hamiltonian flow, while $g_\varphi(q)a_t$ represents the action-dependent control contribution to the momentum-like dynamics.

For an autonomous Hamiltonian system without external control, $a_t=0$, the Hamiltonian is constant along the continuous trajectory. With the control term in Equation~\ref{eq:controlled_hamiltonian}, its time derivative becomes

\begin{equation}
\label{eq:controlled_energy_change}
    \frac{dH_\varphi}{dt}
    =
    \nabla_p H_\varphi(q,p)^\top g_\varphi(q)a_t .
\end{equation}
Thus, Hamiltonian variation separates the effect of external control from the internally conservative Hamiltonian flow~\citep{toth2019hamiltonian}. HTC$^2$ does not directly maximize this variation; instead, it uses the corresponding multi-step quantity as a bounded modulation of temporal future discovery.

\rwhead{Finite-Horizon Future Distribution}
Temporal contrastive learning requires a distribution over future phase states reached from the current state--action pair~\citep{ho2016generative,torabi2018generative,eysenbach2020c}. Let $\mathcal{K}=\{1,\ldots,K\}$ denote the set of admissible future horizons. We define the finite-horizon future distribution as
\begin{equation}
\label{eq:finite_future_occupancy}
    p_{\mathcal{K}}(z_f\mid z_t,a_t)
    =
    \sum_{k=1}^{K}
    \omega_k\,
    p_{\mathcal{T}}
    \left(
        z_{t+k}=z_f
        \mid
        z_t,a_t
    \right),
\end{equation}
where $p_{\mathcal{T}}$ denotes the trajectory distribution and $\omega_k$ is the probability of sampling horizon $k$.

HTC$^2$ uses a truncated geometric horizon distribution,
\begin{equation}
\label{eq:horizon_distribution}
    \omega_k
    =
    \frac{
        (1-\gamma_{\mathrm{T}})
        \gamma_{\mathrm{T}}^{k-1}
    }{
        1-\gamma_{\mathrm{T}}^{K}
    },
    \qquad
    k\in\mathcal{K},
\end{equation}

which satisfies $\sum_{k=1}^{K}\omega_k=1$. The parameter $\gamma_{\mathrm{T}}\in(0,1)$ controls preference for nearer futures while retaining explicitly multi-step positive pairs. Each pair is constrained to lie within the same episode.

\rwhead{Hamiltonian Phase-Space World Model}
HTC$^2$ uses a recurrent Hamiltonian latent world model to construct the phase representation on which temporal discovery is performed. Given an observation $x_t$ and recurrent deterministic state $h_t$, the posterior model infers
\begin{equation}
\label{eq:posterior_phase}
    q_\varphi(z_t\mid h_t,x_t),
    \qquad
    z_t=(q_t,p_t).
\end{equation}
The stochastic phase state is divided into coordinate-like and momentum-like components. In implementation, both components are represented by categorical stochastic variables; their posterior expectations are used whenever continuous phase values are required by the Hamiltonian transition or the temporal critic.

The deterministic state is updated recurrently from the previous phase state and action. The controlled Hamiltonian dynamics in Equation~\ref{eq:controlled_hamiltonian} then provide a physics-structured phase prediction, which is used together with the recurrent context to parameterize the next stochastic prior:
\begin{equation}
\label{eq:ham_prior}
    p_\varphi
    \left(
        z_{t+1}
        \mid
        h_{t+1},z_t,a_t
    \right).
\end{equation}
The Hamiltonian transition is integrated using a leapfrog step, while a learned residual prior retains sufficient flexibility for high-dimensional visual control. Detailed transition equations are given in Appendix~\ref{sec:implementation}.

The world model is optimized with
\begin{equation}
\label{eq:wm_loss}
    \mathcal{L}_{\mathrm{WM}}
    =
    \mathcal{L}_{\mathrm{rec}}
    +
    \alpha\mathcal{L}_{\mathrm{dyn}}
    +
    (1-\alpha)\mathcal{L}_{\mathrm{rep}}
    +
    \mathcal{L}_{\mathrm{inv}}
    +
    \lambda_{\mathrm{phys}}
    \mathcal{L}_{\mathrm{phys}} .
\end{equation}
Here, $\mathcal{L}_{\mathrm{rec}}$ grounds the posterior latent state in the observation space; $\mathcal{L}_{\mathrm{dyn}}$ and $\mathcal{L}_{\mathrm{rep}}$ provide balanced prior--posterior regularization; $\mathcal{L}_{\mathrm{inv}}$ preserves action-relevant information; and $\mathcal{L}_{\mathrm{phys}}$ aligns the Hamiltonian prediction with the posterior phase transition.

For HTC$^2$, the world model provides posterior phase trajectories $(z_t,z_{t+1},\ldots)$ for temporal pair construction and the learned Hamiltonian $H_\varphi(z_t)$ for evaluating multi-step Hamiltonian variation. Temporal positives are formed from actually reached same-episode posterior futures.

\subsection{Hamiltonian Temporal Contrastive Curiosity}
\label{sec:htc2}
HTC$^2$ uses horizon-conditioned temporal contrastive mismatch as its primary curiosity signal and bounded multi-step Hamiltonian variation to modulate this signal.

\rwhead{Horizon-Conditioned Temporal Contrastive Critic}
Contrastive representation learning brings positive pairs closer while separating them from negative samples \citep{chopra2005learning,oord2018representation,wang2020understanding,saunshi2019theoretical,hassani2020contrastive}. In HTC$^2$, the anchor is $(z_t,a_t,k)$, the reached same-episode future $z_{t+k}$ forms the positive pair, and other minibatch futures serve as negatives. The horizon $k\sim\omega_k$ is sampled according to Equation~\ref{eq:horizon_distribution}, enabling the critic to learn future relations over multiple temporal scales.

The critic consists of an anchor encoder and a future encoder:
\begin{equation}
\label{eq:htc_encoders}
    u_{t,k}=\phi_\theta(z_t,a_t,e_k),
    \qquad
    v_{t+k}=\psi_\theta(z_{t+k}).
\end{equation}
where $e_k$ is a learnable horizon embedding.

The temporal score is defined by the negative representation distance:
\begin{equation}
\label{eq:htc_score}
    C_\theta((z_t,a_t,k),z_f)
    =
    -
    \left\|
        \phi_\theta(z_t,a_t,e_k)
        -
        \psi_\theta(z_f)
    \right\|_2 .
\end{equation}
Given a minibatch of $B$ same-episode positive pairs, we optimize the temporal critic with the InfoNCE objective~\citep{oord2018representation}:

\begin{equation}
\label{eq:tc_infonce}
    \mathcal{L}_{\mathrm{TC}}
    =
    -
    \frac{1}{B}
    \sum_{i=1}^{B}
    \log
    \frac{
        \exp
        \left(
        C_\theta(
        (z_{t_i}^i,a_{t_i}^i,k_i),
        z_{t_i+k_i}^i
        )/\tau
        \right)
    }{
        \sum_{j=1}^{B}
        \exp
        \left(
        C_\theta(
        (z_{t_i}^i,a_{t_i}^i,k_i),
        z_{t_j+k_j}^j
        )/\tau
        \right)
    } .
\end{equation}
The matched future forms the positive pair, whereas the remaining minibatch futures act as negatives. The temperature $\tau$ controls the sharpness of the contrastive classification.

The policy does not directly optimize $\mathcal{L}_{\mathrm{TC}}$. Instead, HTC$^2$ extracts the positive-pair distance
\begin{equation}
\label{eq:temporal_distance}
    D_\theta(t,k)
    =
    \left\|
        \phi_\theta(z_t,a_t,e_k)
        -
        \psi_\theta(z_{t+k})
    \right\|_2 .
\end{equation}
A large $D_\theta(t,k)$ indicates that the reached future remains poorly aligned under the current temporal representation. As these transitions are revisited, the temporal critic learns to better organize them through Equation~\ref{eq:tc_infonce}, making the temporal mismatch an adaptive exploration signal.

\rwhead{Hamiltonian-Modulated Temporal Curiosity}
For the same sampled future horizon $k$, HTC$^2$ evaluates the multi-step Hamiltonian variation
\begin{equation}
\label{eq:multi_step_delta_h}
    \Delta H_{t,k}
    =
    \left|
        H_\varphi(z_{t+k})
        -
        H_\varphi(z_t)
    \right|.
\end{equation}
Rather than directly maximizing $\Delta H_{t,k}$, HTC$^2$ converts it into the bounded modulation
\begin{equation}
\label{eq:energy_gate}
    G_H(t,k)
    =
    1
    +
    \alpha_H
    \tanh
    \left(
        \operatorname{clip}
        \left(
            \frac{
                \Delta H_{t,k}
            }{
                \sigma_H+\epsilon
            },
            0,
            c_H
        \right)
    \right),
\end{equation}
where $\sigma_H$ denotes a running scale estimate, $\epsilon$ provides numerical stability, $\alpha_H$ controls the modulation strength, and $c_H$ is the clipping threshold. Since the baseline value of the gate is one and its amplification is bounded, Hamiltonian variation reweights rather than replaces temporal discovery.

The raw HTC$^2$ reward is
\begin{equation}
\label{eq:htc2_reward}
    r_{\mathrm{HTC^2}}^{\mathrm{raw}}(t)
    =
    D_\theta(t,k)G_H(t,k)
    -
    \lambda_{\mathrm{sm}}
    \left\|
        a_t-a_{t-1}
    \right\|_2^2 ,
\end{equation}
where $\lambda_{\mathrm{sm}}$ controls an action-smoothness penalty. The reward is normalized before being combined with the cold-start signal.

Early in training, both the phase representation and the temporal critic are still being learned. We therefore use normalized latent RND, $\bar r_t^{\mathrm{RND}}$, as a warm-start exploration signal. Let $n$ denote the global environment step. The transition weight is
\begin{equation}
\label{eq:rnd_schedule}
    w_n
    =
    \operatorname{clip}
    \left(
        \frac{
            n-T_{\mathrm{warm}}
        }{
            T_{\mathrm{anneal}}
        },
        0,
        1
    \right).
\end{equation}
The final intrinsic reward is
\begin{equation}
\label{eq:rnd_warm_start}
    r_t^{\mathrm{int}}
    =
    (1-w_n)\bar r_t^{\mathrm{RND}}
    +
    w_n\bar r_t^{\mathrm{HTC^2}},
\end{equation}
where $\bar r_t^{\mathrm{HTC^2}}$ denotes the normalized HTC$^2$ signal. The reward used by the policy optimizer is therefore
\begin{equation}
\label{eq:augmented_reward}
    \tilde r_t
    =
    r_t^{\mathrm{ext}}
    +
    \beta r_t^{\mathrm{int}},
\end{equation}
where $\beta$ is the global intrinsic-reward scale. Optimizer-specific training procedures for SAC and PPO are provided in Appendix~\ref{supp:training_pseudocode}.

%% file: main/4_experiments.tex
\section{Experiments}
\label{sec:experiments}

We evaluate HTC$^2$ on the DeepMind Control Suite (DMC)~\citep{tunyasuvunakool2020dm_control} through five complementary studies. First, we compare control performance under SAC and PPO to assess whether HTC$^2$ improves exploration across policy optimizers. Second, we measure physical-state coverage to examine how broadly each method explores the underlying state space. Third, we evaluate zero-shot retention under parametric dynamics shifts to test policy robustness. Fourth, component ablations isolate the contribution of the main HTC$^2$ design choices. Finally, Hamiltonian diagnostics examine whether the learned phase representation exhibits the intended energy and latent structure. Additional results and diagnostics are reported in Appendix~\ref{app:additional_experiments}.

\subsection{Experimental Setup}
\label{subsec:exp_setup}

\paragraph{Benchmarks, baselines, and evaluation.}
We evaluate HTC$^2$ on six tasks: \texttt{walker\_walk}, \texttt{walker\_run}, \texttt{walker\_stand}, \texttt{hopper\_hop}, \texttt{hopper\_stand}, and \texttt{cheetah\_run}. HTC$^2$ is evaluated with both SAC~\citep{haarnoja2018soft} and PPO~\citep{schulman2017proximal}, and compared with four representative exploration methods: ICM~\citep{pathak2017curiosity}, RIDE~\citep{raileanu2020ride}, E3B~\citep{henaff2022exploration}, and Plan2Explore~\citep{sekar2020planning}. Within each policy optimizer, all methods use the same training budget and evaluation settings. Results are reported as mean $\pm$ standard deviation, with the number of random seeds specified for each experiment. Detailed numerical results and complete learning curves are provided in Appendix~\ref{app:dmc_detailed_results}.

\subsection{Control Performance on DMC}
\label{subsec:control_performance}

\begin{figure}[t]
    \centering
    \includegraphics[width=\linewidth]{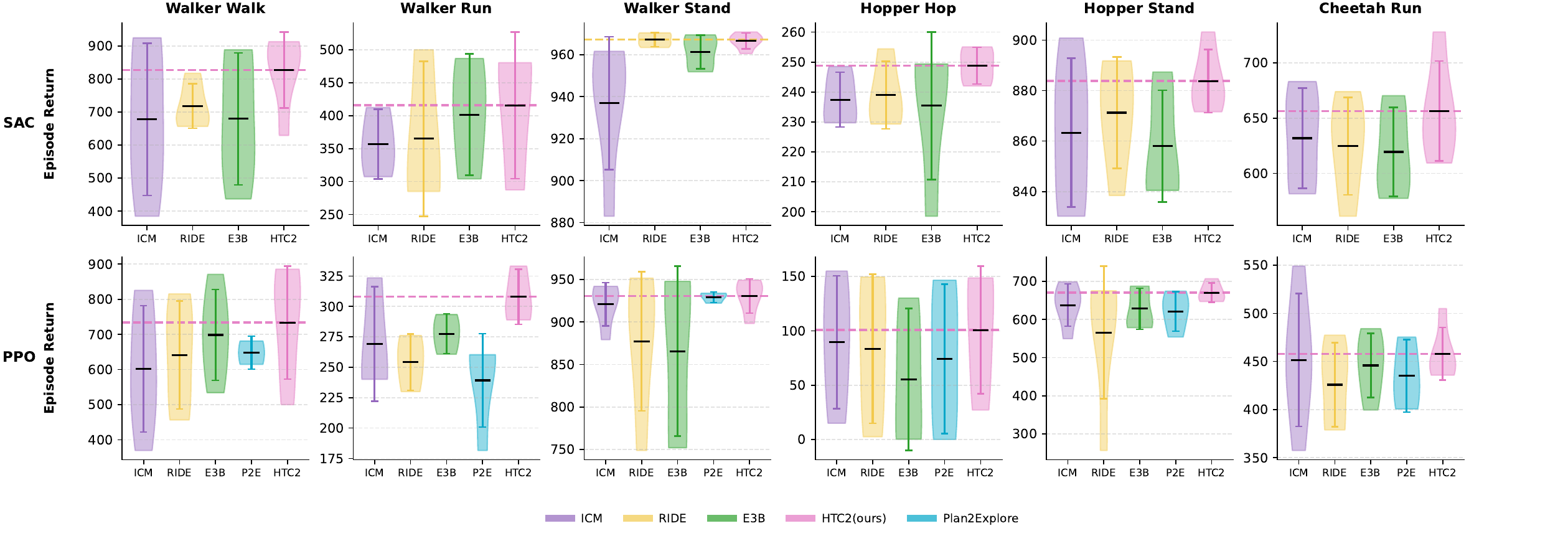}
    \caption{\textbf{Control performance of HTC$^2$ on six DMC tasks with SAC and PPO.} The figure summarizes episode-return performance across five random seeds for ICM, RIDE, E3B, Plan2Explore, and HTC$^2$. Numerical results are reported as mean $\pm$ standard deviation in Appendix~\ref{app:dmc_detailed_results}.}
    \label{fig:dmc_violin_all}
\end{figure}

Figure~\ref{fig:dmc_violin_all} shows that HTC$^2$ maintains strong control performance under both SAC and PPO in comparison with ICM, RIDE, E3B, and Plan2Explore. Across the twelve tasks, HTC$^2$ achieves the highest mean final return in eleven settings. Under SAC, HTC$^2$ ranks first on five of the six tasks and performs nearly identically to the strongest baseline on \texttt{walker\_stand}; under PPO, it achieves the highest mean return on all six tasks. The consistency across the two optimization regimes indicates that the performance of HTC$^2$ is not specific to either off-policy replay or on-policy updates.

The gains are particularly visible on tasks requiring coordinated multi-step transitions, such as \texttt{walker\_walk} and \texttt{hopper\_hop}. This trend is consistent with the design of HTC$^2$: exploration is driven by the mismatch between the current state--action representation and its reached multi-step future, while Hamiltonian variation only provides a bounded modulation of this temporal signal. Detailed per-task returns and complete SAC/PPO learning curves are reported in Appendix~\ref{app:dmc_detailed_results}.

%==================================================================================================

%==================================================================================================
\subsection{Physical-State Exploration Coverage}
\label{subsec:coverage}

We evaluate physical-state coverage on three DMC tasks using low-dimensional privileged simulator states, which are used only for evaluation and are not provided to the policy or intrinsic-reward module. For environment $\mathcal{E}$, let $s_t^{\mathrm{phy}}$ denote the task-specific physical state and $b_{\mathcal{E}}(\cdot)$ its fixed grid discretization. The cumulative coverage at training step $t$ is
\begin{equation}
    \mathcal{C}_t
    =
    \left|
        \left\{
            b_{\mathcal{E}}(s_\tau^{\mathrm{phy}})
            \;:\;
            0\leq \tau\leq t
        \right\}
    \right|,
    \label{eq:state_coverage}
\end{equation}
i.e., the number of unique discretized physical states visited.

For \texttt{cartpole\_swingup}, we use the cart position and wrapped pole angle, $(x_{\mathrm{cart}},\theta_{\mathrm{pole}})$. For \texttt{ball\_in\_cup\_catch}, we use the ball position relative to the cup, $(x_{\mathrm{ball}}-x_{\mathrm{cup}}, z_{\mathrm{ball}}-z_{\mathrm{cup}})$. For \texttt{finger\_spin}, we use the wrapped object angle and clipped angular velocity, $(\theta_{\mathrm{obj}},\dot{\theta}_{\mathrm{obj}})$. Discretization details are provided in Appendix~\ref{app:coverage_details}.

\begin{figure}[t]
    \centering
    \includegraphics[width=\linewidth]
    {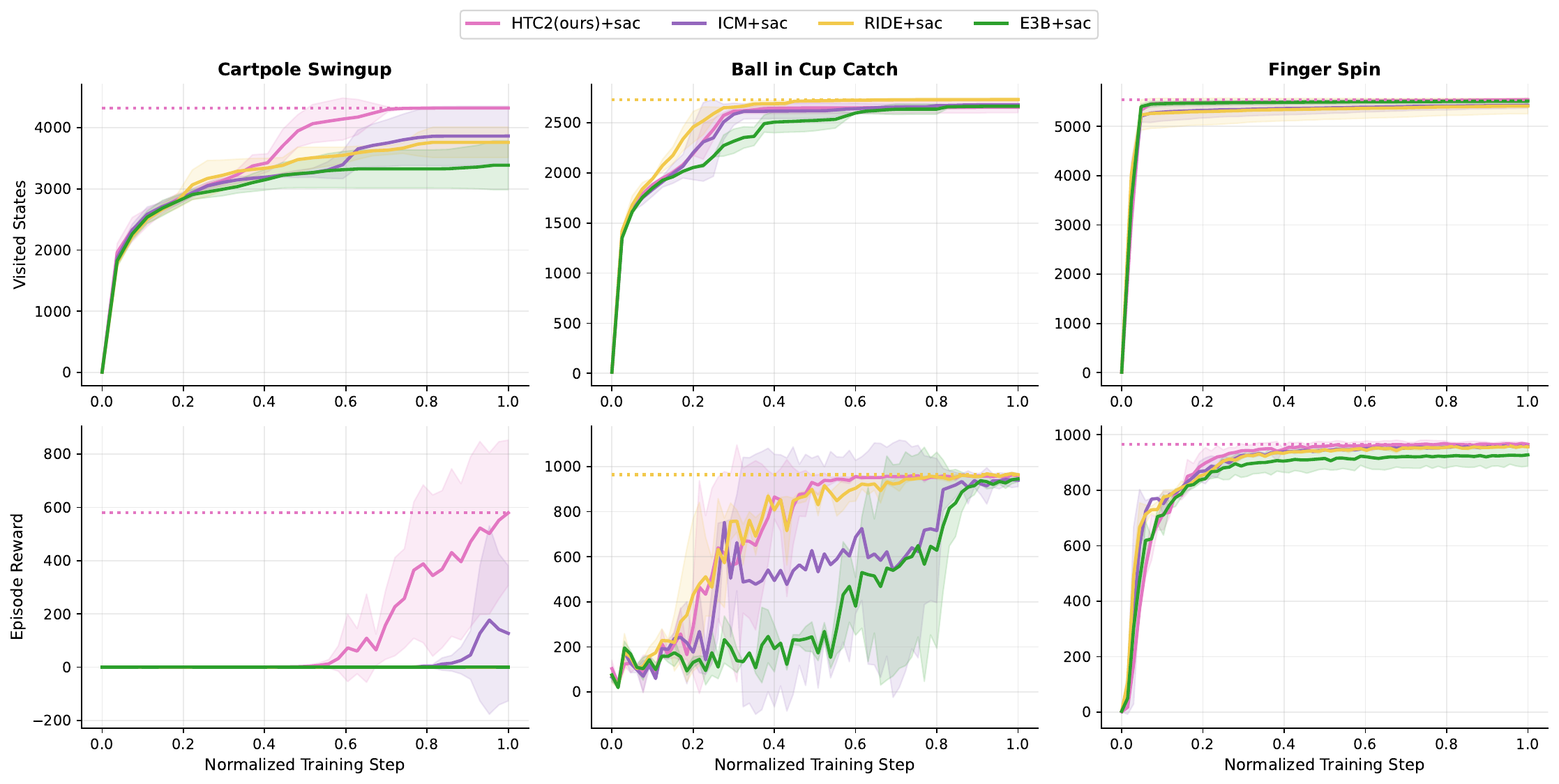}
    \caption{
    \textbf{Physical-state coverage and episode reward during training.}
    Top: cumulative unique discretized states; bottom: episode reward on
    \texttt{cartpole\_swingup}, \texttt{ball\_in\_cup\_catch}, and
    \texttt{finger\_spin}. Curves show mean $\pm$ one standard deviation
    over four random seeds.
    }
    \label{fig:coverage}
\end{figure}

Figure~\ref{fig:coverage} shows distinct coverage patterns across the three environments. On \texttt{cartpole\_swingup}, HTC$^2$ continues expanding its visited-state set after the other methods begin to saturate, reaching the highest final coverage and mean episode reward. On \texttt{ball\_in\_cup\_catch}, RIDE attains the highest coverage, while HTC$^2$ ranks second in coverage AUC, final reward, and reward AUC. On \texttt{finger\_spin}, HTC$^2$ achieves the highest final coverage and final reward, with a coverage AUC close to E3B.

Overall, HTC$^2$ exhibits strong physical-state coverage on \texttt{cartpole\_swingup} and \texttt{finger\_spin}, while remaining competitive on \texttt{ball\_in\_cup\_catch}. Full coverage and reward statistics are reported in Appendix~\ref{app:coverage_details}.

\subsection{Hamiltonian Mechanism Validation}
\label{subsec:hamiltonian_diagnostics}

Beyond policy learning and exploration coverage, we examine whether the learned Hamiltonian phase representation exhibits the structure used by HTC$^2$. Specifically, we test whether $H_\varphi(q,p)$ remains stable under passive evolution, responds to policy-induced control, and varies coherently over the coordinate-like and momentum-like latent subspaces.

\begin{figure}[t]
    \centering
    \includegraphics[width=0.96\linewidth]
    {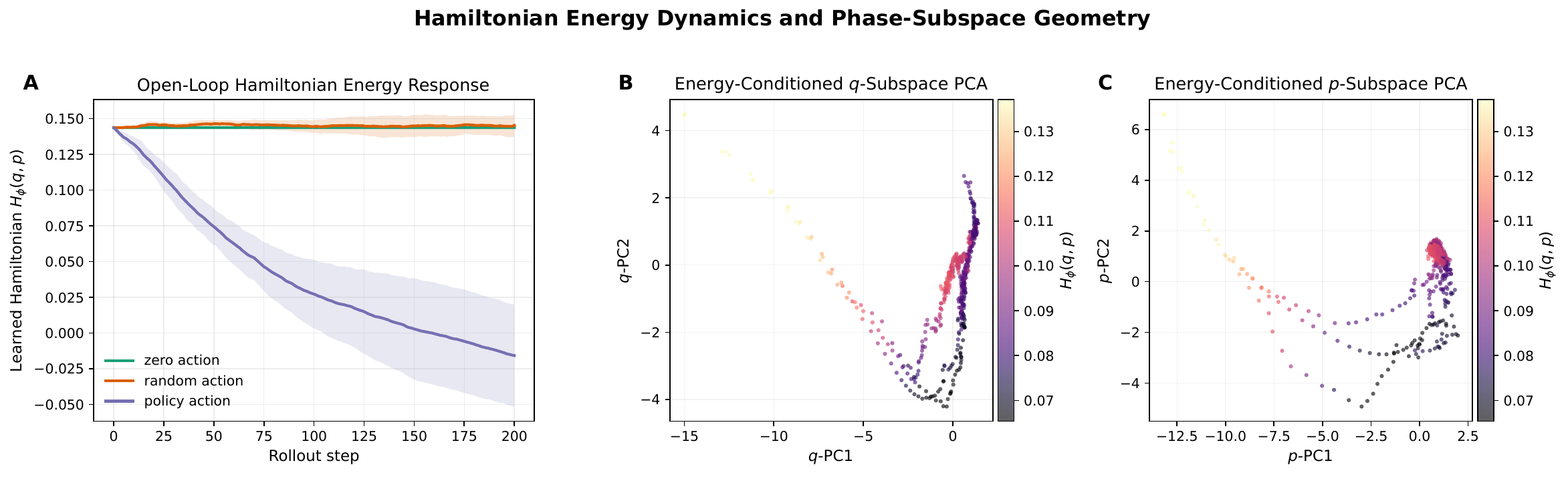}
    \caption{
    \textbf{Hamiltonian energy dynamics and phase-subspace geometry.}
    \textbf{A}: open-loop learned Hamiltonian evolution under zero,
    random, and policy actions.
    \textbf{B--C}: PCA projections of the coordinate-like latent $q$
    and momentum-like latent $p$, colored by $H_\varphi(q,p)$.
    }
    \label{fig:hamiltonian_phase_diagnostics}
\end{figure}

\noindent\textbf{(A) Controlled Hamiltonian energy response.}
Figure~\ref{fig:hamiltonian_phase_diagnostics}A compares the learned Hamiltonian under three action regimes. Under zero action, $H_\varphi(q,p)$ remains nearly flat over the rollout horizon, consistent with approximate energy preservation under autonomous Hamiltonian evolution. Random actions produce similarly bounded deviations around the zero-action trajectory. In contrast, policy actions induce a clear and sustained energy displacement with greater long-horizon variability, showing that the learned Hamiltonian responds differently to structured control.

\noindent\textbf{(B) Energy organization in the $q$ subspace.}
Figure~\ref{fig:hamiltonian_phase_diagnostics}B projects the coordinate-like latent $q$ into two principal components and colors each state by $H_\varphi(q,p)$. The projected trajectory exhibits a clear energy-ordered structure: high-energy states are concentrated near one end of the curve, while lower-energy states occupy a different region of the projected subspace. This smooth variation indicates that the learned Hamiltonian is systematically coupled to the $q$ component of the phase representation. Therefore, the energy term used by HTC$^2$ is computed over an organized phase subspace rather than over an unstructured latent cloud.

\noindent\textbf{(C) Energy organization in the $p$ subspace.}
Figure~\ref{fig:hamiltonian_phase_diagnostics}C shows the corresponding projection for the momentum-like latent $p$. The learned energy again varies smoothly across the projected trajectory, with a compact low-energy region and a separate high-energy branch. This indicates that $H_\varphi(q,p)$ depends on both components of the learned phase state. Together, the $q$ and $p$ visualizations support the design of HTC$^2$: temporal future discovery is performed in a Hamiltonian phase space whose energy varies coherently over the learned latent dynamics, and whose controlled evolution is modulated by the policy actions.

We further analyze the relation between the $q$ and $p$ subspaces in Appendix~\ref{app:latent_geometry}. The dimension-wise coupling maps show that $q$ and $p$ are dynamically related but not uniformly coupled: only selected cross-dimensional pairs exhibit strong correlations. This supports a non-collapsed phase representation in which coordinate-like and momentum-like variables share the same trajectory dynamics while preserving distinct representational roles.
%================================================================================

%================================================================================

\subsection{Parametric OOD Robustness}
\label{subsec:ood}

We further evaluate the zero-shot robustness of SAC policies trained with HTC$^2$ under parametric dynamics shifts. The final policies trained on the original DMC environments are evaluated on variants of \texttt{walker\_walk}, \texttt{hopper\_stand}, and \texttt{cheetah\_run} with modified gravity ($1.5\times$), friction ($2.0\times$), mass ($1.5\times$), or damping ($2.0\times$). We report return retention, defined as the OOD return divided by the corresponding in-distribution return for the same method and seed.

\begin{figure*}[t]
    \centering
    \includegraphics[width=\textwidth]
    {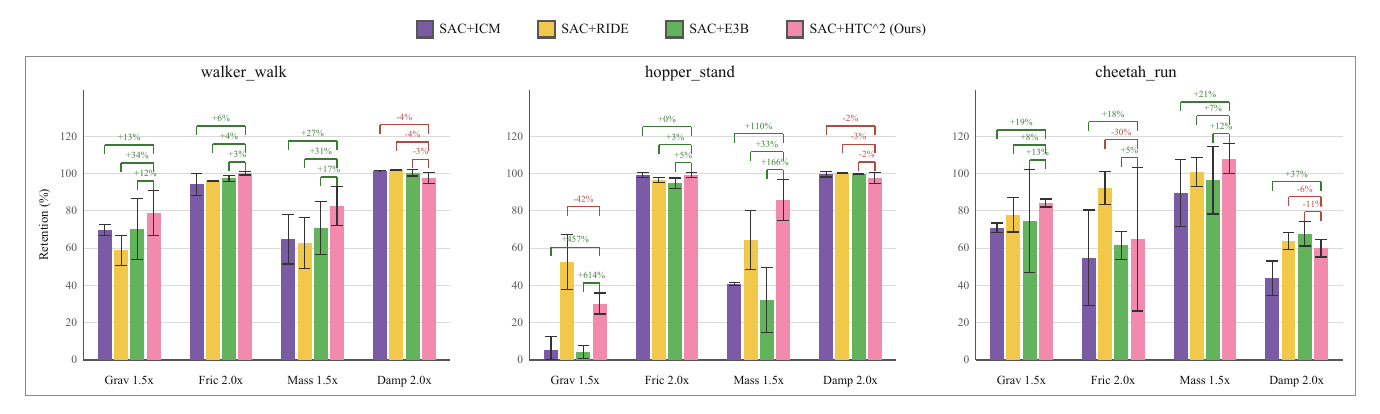}
    \caption{
    \textbf{Zero-shot policy retention under parametric physical shifts.}
    Bars and error bars show mean $\pm$ standard deviation over four
    random seeds. Higher values indicate greater return retention.
    }
    \label{fig:parametric_ood}
\end{figure*}

Figure~\ref{fig:parametric_ood} shows that HTC$^2$ maintains high retention under several physical shifts. On \texttt{walker\_walk}, HTC$^2$ achieves the highest retention under gravity, friction, and mass perturbations, with an average retention of $89.7\%$ across the four variants. On \texttt{hopper\_stand}, it performs strongly under the mass shift and remains competitive under friction and damping, while RIDE obtains the highest retention under the gravity shift. On \texttt{cheetah\_run}, HTC$^2$ achieves the highest retention under gravity and mass perturbations, with mass-shift retention exceeding $100\%$.

Across all three tasks and four parameter shifts, HTC$^2$ obtains the highest overall average retention ($82.3\%$), compared with RIDE ($80.6\%$), E3B ($72.5\%$), and ICM ($69.4\%$). Detailed per-shift results are reported in Appendix~\ref{app:ood_details}.

\vspace*{-0.6em}
\subsection{Ablation Study}
\label{subsec:ablation}

We ablate the main components of HTC$^2$ on \texttt{walker\_walk}, \texttt{hopper\_hop}, and \texttt{cheetah\_run}. Specifically, w/o $\mathcal{L}_{\mathrm{phys}}$ removes the physical-consistency loss; w/o $G_H$ removes the Hamiltonian modulation; one-step $\Delta H$ replaces the temporal contrastive reward with one-step Hamiltonian change; w/o RND removes the warm-start schedule; and w/o repr. norm disables contrastive representation normalization. All variants use SAC with the same training budget and are evaluated over four random seeds.

\begin{table}[t]
\centering
\captionsetup{skip=3pt}
\caption{\textbf{HTC$^2$ ablations.} Final return (mean $\pm$ std, four seeds).}
\label{tab:htc2_ablation}

\small
\setlength{\tabcolsep}{6pt}
\renewcommand{\arraystretch}{1.05}

\resizebox{\linewidth}{!}{%
\begin{tabular}{lccc}
\toprule
\textbf{Variant} & \textbf{Walker} & \textbf{Hopper} & \textbf{Cheetah} \\
\midrule
w/o $\mathcal{L}_{\mathrm{phys}}$ & $693.166 \pm 203.967$ & $245.416 \pm 14.651$ & $566.209 \pm 159.871$ \\
w/o $G_H$ & $718.941 \pm 16.122$ & $235.203 \pm 1.896$ & $609.108 \pm 16.185$ \\
one-step $\Delta H$ & $761.051 \pm 195.324$ & $1.843 \pm 1.183$ & $510.093 \pm 128.266$ \\
w/o RND & $785.069 \pm 93.809$ & $227.774 \pm 0.719$ & $578.816 \pm 9.910$ \\
w/o repr. norm & $757.758 \pm 114.117$ & $144.466 \pm 85.169$ & $594.697 \pm 16.876$ \\
\midrule
\textbf{HTC$^2$ full} & $\mathbf{814.342 \pm 162.857}$ & $\mathbf{250.865 \pm 9.598}$ & $\mathbf{659.810 \pm 33.525}$ \\
\bottomrule
\end{tabular}%
}
\vspace{-0.5em}
\end{table}

Table~\ref{tab:htc2_ablation} shows that the full HTC$^2$ configuration achieves the highest mean final return on all three tasks. Removing $\mathcal{L}_{\mathrm{phys}}$ or $G_H$ lowers the mean return across all environments, with larger reductions from removing $\mathcal{L}_{\mathrm{phys}}$ on \texttt{walker\_walk} and \texttt{cheetah\_run}. Replacing the temporal contrastive reward with one-step $\Delta H$ produces the largest reduction on \texttt{hopper\_hop}, where the final return drops to $1.843\pm1.183$.

Removing the RND warm start or representation normalization also reduces performance across all three tasks. The reduction from removing representation normalization is particularly pronounced on \texttt{hopper\_hop}, whereas removing the RND warm start produces a larger absolute decrease on \texttt{cheetah\_run}. Across the three tasks, the complete configuration yields the strongest mean final performance. Full learning curves are reported in Appendix~\ref{app:ablation_learning_curves}.

%% file: main/5_conclusion.tex
\section{Conclusion}
\label{sec:conclusion}

We presented \textbf{Hamiltonian Temporal Contrastive Curiosity} (\textbf{HTC}$^2$), an intrinsic exploration method for discovering temporally informative futures in a learned Hamiltonian phase space. HTC$^2$ uses a horizon-conditioned temporal contrastive critic to identify poorly organized multi-step futures and modulates this signal with bounded Hamiltonian variation to emphasize structured changes in the learned phase dynamics. Across DeepMind Control Suite tasks, HTC$^2$ shows strong control performance with both SAC and PPO, competitive physical-state coverage, and robust retention under parametric dynamics shifts. Ablations show performance degradation when removing physical consistency, Hamiltonian modulation, warm start, or representation normalization, while the single-step Hamiltonian-change reward performs substantially worse on \texttt{hopper\_hop}. Hamiltonian diagnostics further reveal energy-organized phase representations with distinct responses to policy-induced control. Overall, these results characterize HTC$^2$ as a physics-structured approach to long-horizon exploration in physical control.

%% file: main/6_appendix.tex
\appendix

\section{Detailed Methodology}
\label{sec:method_details}

This section provides additional derivations for the controlled Hamiltonian relations introduced in Equation~\ref{eq:controlled_hamiltonian} and~\ref{eq:controlled_energy_change}, followed by the leapfrog discretization used by the Hamiltonian transition core. We then derive a continuous-time interpretation of the multi-step Hamiltonian variation defined in Equation~\ref{eq:multi_step_delta_h}.

\subsection{Controlled Hamiltonian Dynamics and Energy Balance}
\label{subsec:controlled_hamiltonian_details}

\paragraph{Autonomous Hamiltonian conservation.}
Starting from the controlled dynamics in Equation~\ref{eq:controlled_hamiltonian}, consider first the autonomous case $a_t=0$. For a Hamiltonian with no explicit time dependence, its derivative along the trajectory is
\begin{equation}
\begin{aligned}
    \frac{dH_\varphi}{dt}
    &=
    \left(
        \frac{\partial H_\varphi}{\partial q}
    \right)^\top
    \dot q
    +
    \left(
        \frac{\partial H_\varphi}{\partial p}
    \right)^\top
    \dot p
    \\
    &=
    \left(
        \frac{\partial H_\varphi}{\partial q}
    \right)^\top
    \frac{\partial H_\varphi}{\partial p}
    -
    \left(
        \frac{\partial H_\varphi}{\partial p}
    \right)^\top
    \frac{\partial H_\varphi}{\partial q}
    \\
    &=0.
\end{aligned}
\label{eq:detail_autonomous_conservation}
\end{equation}
This is the standard conservation relation of an autonomous Hamiltonian system. In classical mechanics, it is associated with time-translation invariance through Noether's theorem: in the absence of explicit time dependence and external forcing, the Hamiltonian remains constant along the continuous trajectory.

\paragraph{Energy balance under control.}
Restoring the action-dependent term in Equation~\ref{eq:controlled_hamiltonian} gives
\begin{equation}
\begin{aligned}
    \frac{dH_\varphi}{dt}
    &=
    \left(
        \frac{\partial H_\varphi}{\partial q}
    \right)^\top
    \frac{\partial H_\varphi}{\partial p}
    \\
    &\quad+
    \left(
        \frac{\partial H_\varphi}{\partial p}
    \right)^\top
    \left(
        -
        \frac{\partial H_\varphi}{\partial q}
        +
        g_\varphi(q)a_t
    \right)
    \\
    &=
    \left(
        \frac{\partial H_\varphi}{\partial p}
    \right)^\top
    g_\varphi(q)a_t .
\end{aligned}
\label{eq:detail_control_energy_derivation}
\end{equation}
Using $\dot q=\partial H_\varphi/\partial p$, the last term can equivalently be written as $\dot q^\top g_\varphi(q)a_t$. Thus, Equation~\ref{eq:detail_control_energy_derivation} recovers the controlled energy-balance relation in Equation~\ref{eq:controlled_energy_change}.

\subsection{Symplectic Integration of the Controlled Dynamics}
\label{subsec:symplectic_integration_details}

The controlled continuous-time dynamics in Equation~\ref{eq:controlled_hamiltonian} must be discretized to construct the Hamiltonian transition prior. Our implementation uses a leapfrog integrator with step size $\Delta t=0.1$, with $a_t$ held fixed over one integration step.

The update follows the kick--drift--kick structure. Starting from $(q_t,p_t)$, the first momentum half-step is
\begin{equation}
\begin{aligned}
    p_{t+\frac{1}{2}}
    =
    p_t
    -
    \frac{\Delta t}{2}
    \frac{\partial H_\varphi}{\partial q}
    (q_t,p_t)
    +
    \frac{\Delta t}{2}
    g_\varphi(q_t)a_t.
\end{aligned}
\label{eq:detail_leapfrog_p_half}
\end{equation}

The intermediate momentum-like state is then used to advance the coordinate-like variable over a full step:
\begin{equation}
\begin{aligned}
    q_{t+1}
    =
    q_t
    +
    \Delta t
    \frac{\partial H_\varphi}{\partial p}
    \left(
        q_t,
        p_{t+\frac{1}{2}}
    \right).
\end{aligned}
\label{eq:detail_leapfrog_q}
\end{equation}

Finally, the momentum update is completed at the new position:
\begin{equation}
\begin{aligned}
    p_{t+1}
    =
    p_{t+\frac{1}{2}}
    -
    \frac{\Delta t}{2}
    \frac{\partial H_\varphi}{\partial q}
    \left(
        q_{t+1},
        p_{t+\frac{1}{2}}
    \right)
    +
    \frac{\Delta t}{2}
    g_\varphi(q_{t+1})a_t.
\end{aligned}
\label{eq:detail_leapfrog_p}
\end{equation}

The three updates constitute the Hamiltonian physics step used in the world model. Gradients of $H_\varphi$ with respect to $q$ and $p$ are evaluated through automatic differentiation, while the learned control term is incorporated into both momentum half-steps.

For the autonomous Hamiltonian component, leapfrog provides a symplectic discretization and is designed to preserve the geometric structure of Hamiltonian dynamics under repeated integration. Symplectic integration does not imply exact conservation of $H_\varphi$ at every discrete step. The complete world-model prior also contains the learned control contribution and a stochastic residual parameterization; accordingly, the leapfrog step is used as a structure-preserving inductive bias for the Hamiltonian transition core.

% \subsection{Multi-Step Hamiltonian Variation in HTC$^2$}
% \label{subsec:multistep_hamiltonian_details}
\subsection{Temporal Hamiltonian Variation}
\label{subsec:multistep_hamiltonian_details}

The controlled energy-balance relation in
% Equation~\ref{eq:controlled_energy_change}
Equation~\ref{eq:controlled_energy_change} provides a continuous-time interpretation of the multi-step Hamiltonian variation used by HTC$^2$. Integrating it over the interval $[t,t+\tau]$ gives
\begin{equation}
\begin{aligned}
    H_\varphi(z_{t+\tau})
    -
    H_\varphi(z_t)
    =
    \int_t^{t+\tau}
    \left(
        \frac{\partial H_\varphi}{\partial p}(z_s)
    \right)^\top
    g_\varphi(q_s)a_s
    \,ds .
\end{aligned}
\label{eq:detail_multistep_integral}
\end{equation}

For a sampled discrete horizon $k$, HTC$^2$ uses the endpoint magnitude $\Delta H_{t,k}$ defined in Equation~\ref{eq:multi_step_delta_h}. Hence, the multi-step term summarizes the Hamiltonian variation across the reached temporal interval rather than only an adjacent one-step change.

HTC$^2$ does not optimize this quantity independently. The endpoint variation is transformed by the bounded gate in Equation~\ref{eq:energy_gate} and modulates the temporal mismatch $D_\theta(t,k)$ defined in Equation~\ref{eq:temporal_distance}. The resulting intrinsic signal is given in Equation~\ref{eq:htc2_reward}. Thus, temporal future discovery remains the primary exploration signal, while Hamiltonian variation provides a bounded structural modulation over the same reached multi-step transition.

%%==========================================================================================================================

%%==========================================================================================================================

\section{Implementation Details}
\label{sec:implementation}

We provide implementation details of the Hamiltonian latent world model, its optimization objectives, the temporal contrastive critic, and the intrinsic-reward schedule used in HTC$^2$. Algorithms~\ref{alg:htc2_sac} and~\ref{alg:htc2_ppo} summarize the integration of HTC$^2$ with SAC and PPO, respectively.

\subsection{Hamiltonian Latent World Model}
\label{supp:world_model}

At time $t$, the recurrent latent state consists of a deterministic state $h_t$ and a phase representation
\begin{equation}
    z_t=[q_t,p_t],
    \label{eq:impl_phase_state}
\end{equation}
where $q_t$ and $p_t$ denote coordinate-like and momentum-like latent variables. The deterministic state is updated by the recurrent core,
\begin{equation}
    h_{t+1}
    =
    \mathrm{GRU}
    \left(
        F_{\mathrm{rnn}}([q_t,p_t,a_t]),
        h_t
    \right),
    \label{eq:impl_gru}
\end{equation}
where $F_{\mathrm{rnn}}$ denotes the input projection preceding the GRU.

Let
\begin{equation}
    c_t=E_{\varphi}(x_t)
    \label{eq:impl_obs_encoding}
\end{equation}
denote the observation token produced by the visual encoder $E_{\varphi}$. The posterior network predicts categorical distributions over the two phase components:
\begin{align}
    Q_{q,t}
    &=
    q_{\varphi}(q_t\mid h_t,c_t),
    \\
    Q_{p,t}
    &=
    q_{\varphi}(p_t\mid h_t,c_t).
    \label{eq:impl_phase_posterior}
\end{align}
Their continuous posterior expectations are
\begin{equation}
    \mu_{q,t}
    =
    \mathbb{E}_{Q_{q,t}}[q_t],
    \qquad
    \mu_{p,t}
    =
    \mathbb{E}_{Q_{p,t}}[p_t].
    \label{eq:impl_phase_expectation}
\end{equation}
For compactness, $q_t$ and $p_t$ below denote these continuous phase values whenever the Hamiltonian transition or temporal critic is evaluated.

The Hamiltonian transition follows the controlled dynamics in Equation~\ref{eq:controlled_hamiltonian}. The learned input mapping
\begin{equation}
    g_{\varphi}(q_t)
    \in
    \mathbb{R}^{d_p\times d_a}
\end{equation}
maps the action $a_t\in\mathbb{R}^{d_a}$ to the momentum-like control term $g_{\varphi}(q_t)a_t\in\mathbb{R}^{d_p}$. Using the leapfrog updates in Equation~\ref{eq:detail_leapfrog_p_half} and~\ref{eq:detail_leapfrog_p}, one deterministic Hamiltonian step is written compactly as
\begin{equation}
    (\bar q_{t+1},\bar p_{t+1})
    =
    \Phi_{\Delta t}^{\mathrm{LF}}
    \left(
        q_t,p_t,a_t;
        H_{\varphi},g_{\varphi}
    \right),
    \label{eq:impl_physics_step}
\end{equation}
where $\Phi_{\Delta t}^{\mathrm{LF}}$ denotes one controlled leapfrog step and $\Delta t=0.1$.

The continuous prediction is subsequently mapped to categorical prior logits. We denote by $\mathcal{G}_{\sigma_\ell}(\cdot)$ the deterministic Gaussian-shaped logit projection with width parameter $\sigma_\ell$. The prior logits are
\begin{align}
    \ell_{q,t+1}^{\mathrm{prior}}
    &=
    \mathcal{G}_{\sigma_\ell}(\bar q_{t+1})
    +
    R_q(\chi_t),
    \\
    \ell_{p,t+1}^{\mathrm{prior}}
    &=
    \mathcal{G}_{\sigma_\ell}(\bar p_{t+1})
    +
    R_p(\chi_t),
    \label{eq:impl_prior_logits}
\end{align}
where
\begin{equation}
    \chi_t=[h_{t+1},q_t,p_t,a_t].
\end{equation}
For $\nu\in\{q,p\}$, the gated residual branch is
\begin{equation}
    R_{\nu}(\chi)
    =
    s\,
    \sigma\!\left(G_{\nu}(\chi)\right)
    \odot
    \tanh\!\left(F_{\nu}(\chi)\right),
    \label{eq:impl_residual}
\end{equation}
where $F_{\nu}$ and $G_{\nu}$ denote the residual-value and gate networks, respectively, $\sigma(\cdot)$ is the sigmoid function, $\odot$ denotes element-wise multiplication, and $s$ is the residual scale. We use $s=1.0$. The residual output is zero-initialized and the gate is initialized negatively so that the Hamiltonian prediction dominates early training.

The categorical prior distributions used below are obtained as
\begin{align}
    P_{q,t+1}
    &=
    \operatorname{Cat}
    \left(
        \operatorname{softmax}
        (\ell_{q,t+1}^{\mathrm{prior}})
    \right),
    \\
    P_{p,t+1}
    &=
    \operatorname{Cat}
    \left(
        \operatorname{softmax}
        (\ell_{p,t+1}^{\mathrm{prior}})
    \right).
    \label{eq:impl_phase_prior}
\end{align}

\subsection{World Model Loss Functions}
\label{supp:world_model_loss}

The complete world-model objective is defined in Equation~\ref{eq:wm_loss}. We use $\alpha=0.8$ and $\lambda_{\mathrm{phys}}=1.0$. This section specifies the individual loss terms used to optimize the posterior representation and the Hamiltonian dynamics prior.

\paragraph{Reconstruction Loss ($\mathcal{L}_{\mathrm{rec}}$).}
Define the posterior state
\begin{equation}
    s_t^{\mathrm{post}}
    =
    [h_t,\mu_{q,t},\mu_{p,t}].
    \label{eq:impl_posterior_state}
\end{equation}
The decoder $D_{\varphi}$ reconstructs the next observation from the posterior state:
\begin{equation}
    \mathcal{L}_{\mathrm{rec}}
    =
    \mathbb{E}
    \left[
        \left\|
            D_{\varphi}(s_{t+1}^{\mathrm{post}})
            -
            x_{t+1}
        \right\|_2^2
    \right].
    \label{eq:impl_rec_loss}
\end{equation}

\paragraph{Dynamics and Representation Losses ($\mathcal{L}_{\mathrm{dyn}},\mathcal{L}_{\mathrm{rep}}$).}
The dynamics loss updates the prior while stopping gradients through the posterior distributions:
\begin{equation}
\begin{aligned}
    \mathcal{L}_{\mathrm{dyn}}
    =
    \mathbb{E}\Big[
    &
    D_{\mathrm{KL}}
    \left(
        \mathrm{sg}[Q_{q,t+1}]
        \Vert
        P_{q,t+1}
    \right)
    \\
    &+
    D_{\mathrm{KL}}
    \left(
        \mathrm{sg}[Q_{p,t+1}]
        \Vert
        P_{p,t+1}
    \right)
    \Big].
\end{aligned}
\label{eq:impl_dyn_loss}
\end{equation}
Conversely, the representation loss updates the posterior while stopping gradients through the prior:
\begin{equation}
\begin{aligned}
    \mathcal{L}_{\mathrm{rep}}
    =
    \mathbb{E}\Big[
    &
    D_{\mathrm{KL}}
    \left(
        Q_{q,t+1}
        \Vert
        \mathrm{sg}[P_{q,t+1}]
    \right)
    \\
    &+
    D_{\mathrm{KL}}
    \left(
        Q_{p,t+1}
        \Vert
        \mathrm{sg}[P_{p,t+1}]
    \right)
    \Big].
\end{aligned}
\label{eq:impl_rep_loss}
\end{equation}
Here, $\mathrm{sg}[\cdot]$ denotes the stop-gradient operator. In the implementation, the $q$- and $p$-KL terms are summed before applying standard free-nats stabilization and batch averaging.

\paragraph{Inverse-Dynamics Loss ($\mathcal{L}_{\mathrm{inv}}$).}
The inverse-dynamics model predicts the executed action from adjacent posterior states,
\begin{equation}
    \hat a_t
    =
    F_{\mathrm{inv}}
    \left(
        [s_t^{\mathrm{post}},s_{t+1}^{\mathrm{post}}]
    \right),
\end{equation}
and is optimized by
\begin{equation}
    \mathcal{L}_{\mathrm{inv}}
    =
    \mathbb{E}
    \left[
        \|\hat a_t-a_t\|_2^2
    \right].
    \label{eq:impl_inv_loss}
\end{equation}

\paragraph{Hamiltonian Consistency Loss ($\mathcal{L}_{\mathrm{phys}}$).}
The continuous Hamiltonian prediction from Equation~\ref{eq:impl_physics_step} is aligned with the posterior phase expectation:
\begin{equation}
\begin{aligned}
    \mathcal{L}_{\mathrm{phys}}
    =
    \mathbb{E}\Big[
    &
    \|
        \bar q_{t+1}
        -
        \mathrm{sg}[\mu_{q,t+1}]
    \|_2^2
    \\
    &+
    \|
        \bar p_{t+1}
        -
        \mathrm{sg}[\mu_{p,t+1}]
    \|_2^2
    \Big].
\end{aligned}
\label{eq:impl_phys_loss}
\end{equation}

\subsection{Temporal Contrastive Critic}
\label{supp:temporal_critic}

The temporal critic operates on posterior phase trajectories inferred from collected experience. Future horizons are sampled according to the truncated geometric distribution in Equation~\ref{eq:horizon_distribution}, with $K=16$ and $\gamma_{\mathrm{T}}=0.98$ in the reported configuration. Only anchor--future pairs belonging to the same episode are retained.

The anchor and future representations follow Equation~\ref{eq:htc_encoders}. Both representations are $\ell_2$-normalized before evaluating the score in Equation~\ref{eq:htc_score}. The critic uses a representation dimension of $64$ and a horizon-embedding dimension of $16$. The temperature is initialized to $0.1$ and optimized jointly with the critic.

The temporal critic is trained using $\mathcal{L}_{\mathrm{TC}}$ in Equation~\ref{eq:tc_infonce} with a batch size of $256$ and AdamW learning rate $3\times10^{-4}$. Gradients are stopped at the phase-state input in the default configuration, so this objective updates the temporal critic without directly propagating gradients into the Hamiltonian world model.

At reward computation time, the temporal mismatch is evaluated using Equation~\ref{eq:temporal_distance}, while the Hamiltonian modulation and the raw HTC$^2$ reward are computed according to Equation~\ref{eq:energy_gate} and~\ref{eq:htc2_reward}, respectively. The HTC$^2$ reward is then normalized using expected-distance normalization. For $\ell_2$-normalized representations and Euclidean distance, the reference distance is $\sqrt{2}$.

\subsection{Cold Start and Reward Scheduling}
\label{supp:reward_schedule}

The cold-start RND~\citep{burda2018exploration} module operates on the posterior phase representation $z_t$. It consists of a trainable predictor $f_\xi$ and a fixed randomly initialized target network $\bar f$. The raw RND signal is
\begin{equation}
    r_{\mathrm{RND}}^{\mathrm{raw}}(t)
    =
    \left\|
        f_\xi(z_t)-\bar f(z_t)
    \right\|_2^2 .
    \label{eq:impl_rnd_reward}
\end{equation}
The predictor and target are MLPs with hidden dimension $256$ and output dimension $128$. The predictor is optimized with AdamW at learning rate $3\times10^{-4}$.

The RND signal is normalized by running statistics before reward mixing. The RND-to-HTC$^2$ transition follows Equation~\ref{eq:rnd_schedule}] and~\ref{eq:rnd_warm_start}, with $T_{\mathrm{warm}}=50{,}000$ and $T_{\mathrm{anneal}}=200{,}000$ environment steps. The resulting intrinsic reward is combined with the extrinsic reward according to Equation~\ref{eq:augmented_reward}.

\subsection{Hyperparameters}
\label{supp:hyperparameters}

The core exploration hyperparameters are summarized in Table~\ref{tab:intrinsic_hyperparams}, while the policy-optimizer configurations are reported in Table~\ref{tab:policy_hyperparams}.

\begin{table}[t]
\centering
\caption{Core hyperparameters for HTC$^2$ and intrinsic-reward baselines.}
\label{tab:intrinsic_hyperparams}
\scriptsize
\setlength{\tabcolsep}{3.2pt}
\resizebox{\linewidth}{!}{
\begin{tabular}{lc@{\hspace{1.3em}}lccc}
\toprule
\multicolumn{2}{c}{\textbf{HTC$^2$}}
&
\multicolumn{4}{c}{\textbf{Intrinsic-reward baselines}}
\\
\cmidrule(lr){1-2}
\cmidrule(lr){3-6}
Hyperparameter & Value
& Hyperparameter & ICM & E3B & RIDE \\
\midrule
Intrinsic scale $\beta$ & 0.05
& Intrinsic scale $\beta$ & 0.05 & 0.05 & 0.05 \\

Cold-start reward & RND
& Latent dim & 256 & 256 & 256 \\

Warmup steps & 50K
& Reward lr & $10^{-3}$ & $10^{-3}$ & $10^{-3}$ \\

Annealing steps & 200K
& Batch size & 256 & 256 & 256 \\

World-model lr & $3\times10^{-4}$
& Update proportion & 1.0 & 1.0 & 1.0 \\

World-model batch size & 256
& Reward normalization & RMS & RMS & RMS \\

Deterministic state dim & 256
& E3B ridge & -- & 0.1 & -- \\

Stochastic latent dim & 32
& RIDE $k$ & -- & -- & 10 \\

$q$ / $p$ dim & 16 / 16
& Kernel cluster distance & -- & -- & 0.008 \\

Categorical classes & 32
& Kernel epsilon & -- & -- & $10^{-4}$ \\

Sequence chunk & 32
& RIDE $c$ & -- & -- & 0.001 \\

Integrator / $\Delta t$ & Leapfrog / 0.1
& RIDE $s_m$ & -- & -- & 8.0 \\

Logit projection std. & 0.1
& $\kappa$ & 0.0 & 0.0 & 0.0 \\

Residual scale & 1.0
& $\gamma$ & -- & -- & -- \\

KL balance $\alpha$ & 0.8
& & & & \\

Free-nats stabilization & 1.0
& & & & \\

Physics loss weight & 1.0
& & & & \\

Temporal horizon $k$ & 1--16
& & & & \\

Horizon sampling & Geometric ($0.98$)
& & & & \\

Temporal critic lr / batch & $3\times10^{-4}$ / 256
& & & & \\

Critic hidden / repr. dim & 256 / 64
& & & & \\

Horizon embedding dim & 16
& & & & \\

Temperature init. & 0.1 (learned)
& & & & \\

Similarity & Negative $\ell_2$
& & & & \\

Representation normalization & Yes
& & & & \\

Energy gate $\alpha_H/c_H$ & 0.1 / 5.0
& & & & \\

Smoothness $\lambda_{\mathrm{sm}}$ & 0.01
& & & & \\
\bottomrule
\end{tabular}
}
\end{table}

\begin{table}[t]
\centering
\caption{SAC and PPO hyperparameters.}
\label{tab:policy_hyperparams}
\begin{tabular}{lcc}
\toprule
Hyperparameter & SAC & PPO \\
\midrule
Learning rate & $1\times10^{-4}$ & $1.5\times10^{-5}$ \\
Feature dim & 50 & 512 \\
Hidden dim & 1024 & 1024 \\
Batch size & 256 & 256 \\
Discount $\gamma$ & 0.99 & 0.99 \\
Initial random steps & 5000 & -- \\
Replay size & 50000 & -- \\
Initial temperature & 0.1 & -- \\
Target update $\tau$ & 0.005 & -- \\
Actor / target update freq. & 1 / 2 & -- \\
Rollout steps & -- & 256 \\
PPO epochs & -- & 15 \\
Clip range / value clip & -- & 0.2 / 0.1 \\
Entropy coefficient & -- & 0.1 \\
Value coefficient & -- & 0.5 \\
Max gradient norm & -- & 0.5 \\
\bottomrule
\end{tabular}
\end{table}

\subsection{Training Pseudocode}
\label{supp:training_pseudocode}

Algorithms~\ref{alg:htc2_sac} and~\ref{alg:htc2_ppo} summarize the integration of HTC$^2$ with off-policy SAC and on-policy PPO, respectively. Standard SAC and PPO optimization steps are omitted for clarity.

\begin{algorithm}[t]
\caption{HTC$^2$ with Soft Actor--Critic}
\label{alg:htc2_sac}
\begin{algorithmic}[1]

\State \textbf{Params:}
$\pi_\eta$,
$Q_{\omega_1},Q_{\omega_2}$,
$p_\varphi$,
$(\phi_\theta,\psi_\theta)$,
$f_\xi$
\Comment{policy, critics, world model, TC critic, RND}

\State \textbf{Init:}
replay buffer $\mathcal D$,
target critics,
fixed RND target $\bar f$
\Comment{initialization}

\While{not converged}

    \State
    $a_t\sim\pi_\eta(\cdot\mid x_t)$,
    \quad
    $\mathcal D
    \leftarrow
    \mathcal D
    \cup
    \{x_t,a_t,r_t^{\mathrm{ext}},x_{t+1},d_t\}$
    \Comment{environment interaction}

    \State
    $\mathcal B_{\mathrm{seq}}\sim\mathcal D$,
    \quad
    $z_{1:T}
    \leftarrow
    q_\varphi^{\mathrm{post}}
    (\mathcal B_{\mathrm{seq}})$
    \Comment{posterior phase trajectory}

    \State
    $k_i\sim\omega_k$,
    \quad
    $z_i^{+}\leftarrow z_{t_i+k_i}$
    \Comment{Equation~\ref{eq:horizon_distribution}}

    \State
    $\theta
    \leftarrow
    \theta
    -
    \eta_{\mathrm{TC}}
    \nabla_\theta
    \mathcal L_{\mathrm{TC}}$
    \Comment{Equation~\ref{eq:tc_infonce}}

    \State
    $r_i^{\mathrm{HTC^2}}
    \leftarrow
    \operatorname{Norm}
    \left(
        D_\theta(t_i,k_i)
        G_H(t_i,k_i)
        -
        \lambda_{\mathrm{sm}}
        \|a_{t_i}-a_{t_i-1}\|_2^2
    \right)$
    \Comment{Equation~\ref{eq:htc2_reward}}

    \State
    $r_i^{\mathrm{RND}}
    \leftarrow
    \operatorname{Norm}
    \left(
        \|f_\xi(z_{t_i})-\bar f(z_{t_i})\|_2^2
    \right)$
    \Comment{latent RND}

    \State
    $w_n
    \leftarrow
    \operatorname{clip}
    \left(
        \frac{n-T_{\mathrm{warm}}}
             {T_{\mathrm{anneal}}},
        0,1
    \right)$
    \Comment{Equation~\ref{eq:rnd_schedule}}

    \State
    $r_i^{\mathrm{int}}
    \leftarrow
    (1-w_n)r_i^{\mathrm{RND}}
    +
    w_n r_i^{\mathrm{HTC^2}}$
    \Comment{Equation~\ref{eq:rnd_warm_start}}

    \State
    $\tilde r_i
    \leftarrow
    r_i^{\mathrm{ext}}
    +
    \beta r_i^{\mathrm{int}}$
    \Comment{Equation~\ref{eq:augmented_reward}}

    \State
    $\mathcal B_{\mathrm{RL}}
    \leftarrow
    \{(x_{t_i},a_{t_i},\tilde r_i,x_{t_i+1},d_{t_i})\}_{i=1}^{B}$
    \Comment{anchor transitions}

    \State
    $(\eta,\omega_1,\omega_2)
    \leftarrow
    \mathrm{SACUpdate}
    (\mathcal B_{\mathrm{RL}})$
    \Comment{standard SAC update}

    \State
    $\varphi
    \leftarrow
    \varphi
    -
    \eta_{\mathrm{WM}}
    \nabla_\varphi
    \mathcal L_{\mathrm{WM}}$
    \Comment{Equation~\ref{eq:wm_loss}}

    \State
    $\xi
    \leftarrow
    \xi
    -
    \eta_{\mathrm{RND}}
    \nabla_\xi
    \|f_\xi(z)-\bar f(z)\|_2^2$
    \Comment{RND predictor update}

\EndWhile
\end{algorithmic}
\end{algorithm}

\begin{algorithm}[t]
\caption{HTC$^2$ with Proximal Policy Optimization}
\label{alg:htc2_ppo}
\begin{algorithmic}[1]

\State \textbf{Params:}
$\pi_\eta$,
$V_\omega$,
$p_\varphi$,
$(\phi_\theta,\psi_\theta)$,
$f_\xi$
\Comment{policy, value, world model, TC critic, RND}

\State \textbf{Init:}
rollout buffer $\mathcal R$,
fixed RND target $\bar f$
\Comment{initialization}

\For{each rollout iteration}

    \State
    $\mathcal R
    \leftarrow
    \{x_t,a_t,r_t^{\mathrm{ext}},x_{t+1},d_t\}_{t=1}^{T_{\mathrm{roll}}}$
    \Comment{on-policy rollout}

    \State
    $z_{1:T}
    \leftarrow
    q_\varphi^{\mathrm{post}}(\mathcal R)$
    \Comment{posterior phase trajectory}

    \State
    $k_i\sim\omega_k$,
    \quad
    $z_i^{+}\leftarrow z_{t_i+k_i}$
    \Comment{Equation~\ref{eq:horizon_distribution}}

    \State
    $\theta
    \leftarrow
    \theta
    -
    \eta_{\mathrm{TC}}
    \nabla_\theta
    \mathcal L_{\mathrm{TC}}$
    \Comment{Equation~\ref{eq:tc_infonce}}

    \State
    $r_i^{\mathrm{HTC^2}}
    \leftarrow
    \operatorname{Norm}
    \left(
        D_\theta(t_i,k_i)
        G_H(t_i,k_i)
        -
        \lambda_{\mathrm{sm}}
        \|a_{t_i}-a_{t_i-1}\|_2^2
    \right)$
    \Comment{Equation~\ref{eq:htc2_reward}}

    \State
    $r_i^{\mathrm{RND}}
    \leftarrow
    \operatorname{Norm}
    \left(
        \|f_\xi(z_{t_i})-\bar f(z_{t_i})\|_2^2
    \right)$
    \Comment{latent RND}

    \State
    $w_n
    \leftarrow
    \operatorname{clip}
    \left(
        \frac{n-T_{\mathrm{warm}}}
             {T_{\mathrm{anneal}}},
        0,1
    \right)$
    \Comment{Equation~\ref{eq:rnd_schedule}}

    \State
    $r_i^{\mathrm{int}}
    \leftarrow
    (1-w_n)r_i^{\mathrm{RND}}
    +
    w_n r_i^{\mathrm{HTC^2}}$
    \Comment{Equation~\ref{eq:rnd_warm_start}}

    \State
    $\tilde r_i
    \leftarrow
    r_i^{\mathrm{ext}}
    +
    \beta r_i^{\mathrm{int}}$
    \Comment{Equation~\ref{eq:augmented_reward}}

    \State
    $(\hat A_i,R_i)
    \leftarrow
    \mathrm{GAE}
    (\mathcal R,\tilde r,V_\omega)$
    \Comment{advantages and returns}

    \State
    $(\eta,\omega)
    \leftarrow
    \mathrm{PPOUpdate}
    (\mathcal R,\hat A,R)$
    \Comment{standard clipped PPO update}

    \State
    $\varphi
    \leftarrow
    \varphi
    -
    \eta_{\mathrm{WM}}
    \nabla_\varphi
    \mathcal L_{\mathrm{WM}}$
    \Comment{Equation~\ref{eq:wm_loss}}

    \State
    $\xi
    \leftarrow
    \xi
    -
    \eta_{\mathrm{RND}}
    \nabla_\xi
    \|f_\xi(z)-\bar f(z)\|_2^2$
    \Comment{RND predictor update}

\EndFor
\end{algorithmic}
\end{algorithm}

%%==========================================================================================================================

%%==========================================================================================================================
\section{Additional Experiments and Diagnostics}
\label{app:additional_experiments}

\subsection{Detailed DMC Control Results}
\label{app:dmc_detailed_results}

\begin{figure}[t]
    \centering
    \includegraphics[width=\linewidth]{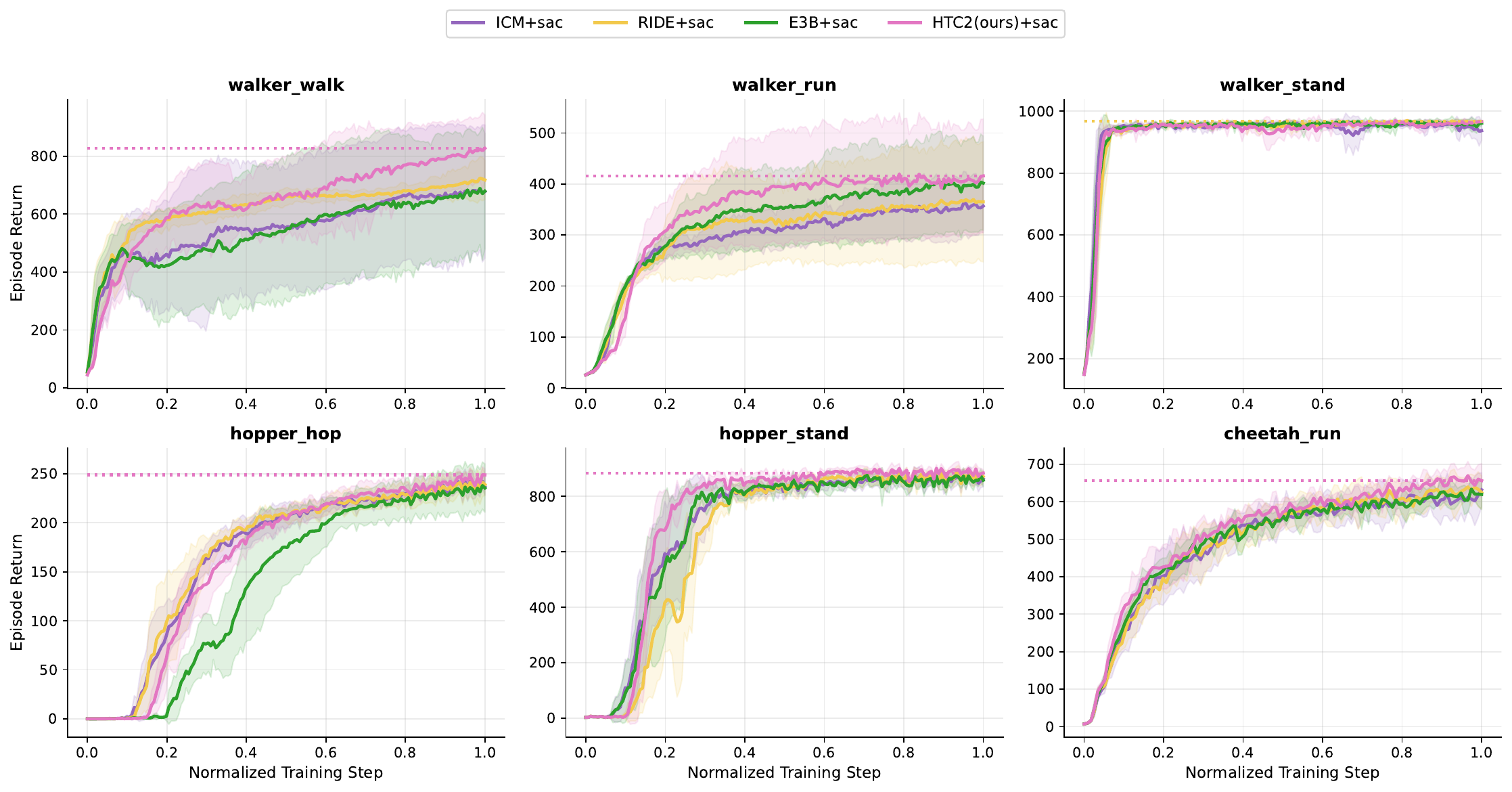}
    \caption{ \textbf{SAC learning curves on six DMC tasks.} Curves show the mean episode return over five random seeds, with shaded regions indicating one standard deviation, for SAC combined with ICM, RIDE, E3B, and HTC$^2$. }
    \label{fig:dmc_sac_all}
\end{figure}

\begin{figure}[t]
    \centering
    \includegraphics[width=\linewidth]{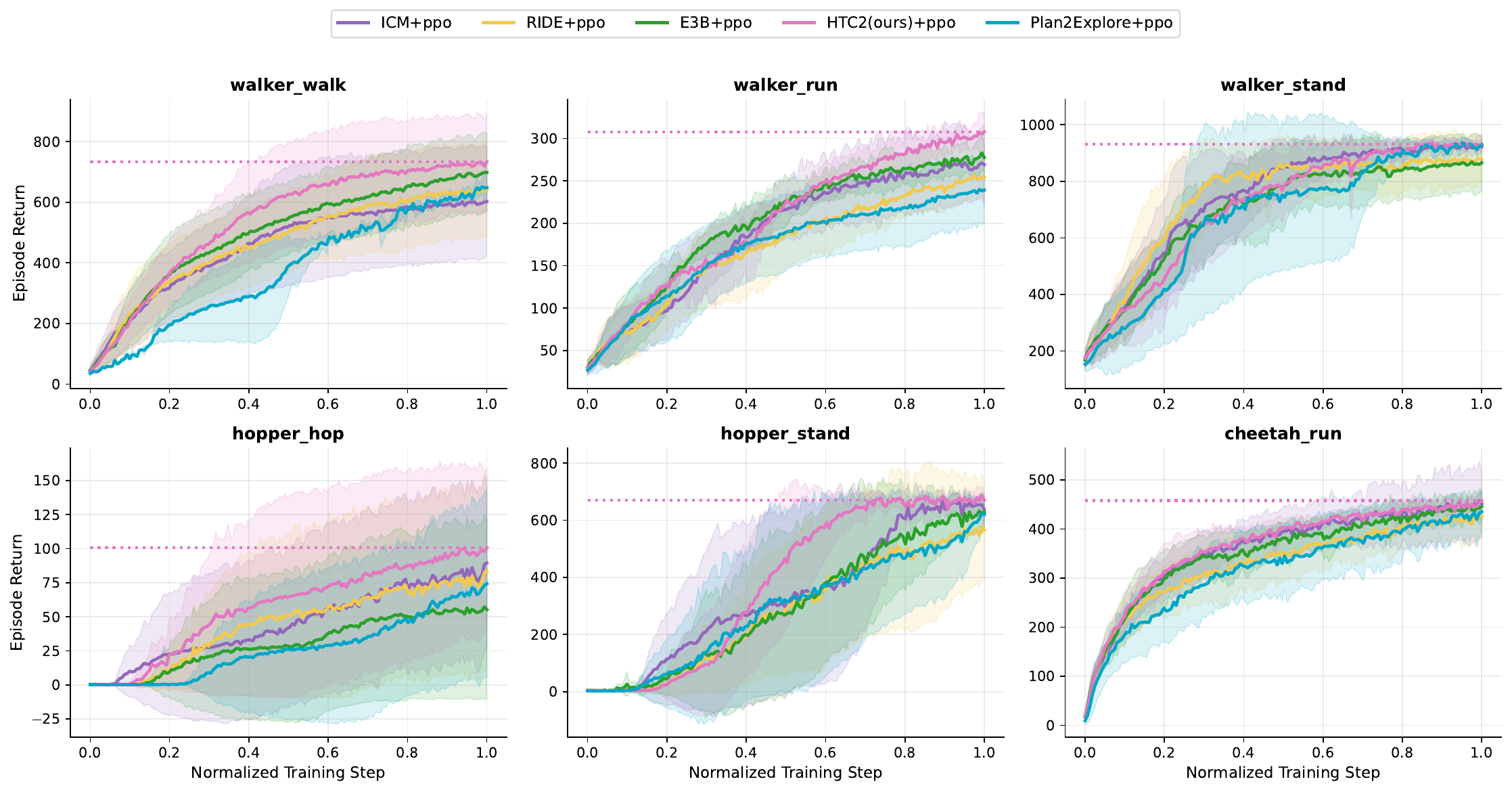}
    \caption{ \textbf{PPO learning curves on six DMC tasks.} Curves show the mean episode return over five random seeds, with shaded regions indicating one standard deviation, for PPO combined with ICM, RIDE, E3B, Plan2Explore, and HTC$^2$. }
    \label{fig:dmc_ppo_all}
\end{figure}

This section provides the complete numerical results and learning curves corresponding to the DMC control benchmark in Section~\ref{subsec:control_performance}. All reported final returns are computed over five random seeds and are shown as mean $\pm$ standard deviation. Table~\ref{tab:dmc_final_return} reports the final control performance for ICM, RIDE, E3B, and HTC$^2$ under SAC and PPO, with Plan2Explore additionally included in the PPO comparison. Figures~\ref{fig:dmc_sac_all} and~\ref{fig:dmc_ppo_all} provide the corresponding SAC and PPO learning curves.

\begin{table}[t]
\centering
\caption{\textbf{Final control performance on six DMC tasks.} Episode return over five random seeds, reported as mean $\pm$ standard deviation. Within each optimizer block and task, red, orange, and yellow denote the first, second, and third-ranked exploration methods, respectively. Higher is better.}
\label{tab:dmc_final_return}
\scriptsize
\setlength{\tabcolsep}{3.0pt}
\renewcommand{\arraystretch}{1.08}
\resizebox{\linewidth}{!}{%
\begin{tabular}{llcccccc}
\toprule
RL & Exploration & Walker Walk & Walker Run & Walker Stand & Hopper Hop & Hopper Stand & Cheetah Run \\
\midrule

\multirow{4}{*}{SAC}
& ICM
& $677.8{\pm}230.2$
& $356.7{\pm}52.9$
& $936.9{\pm}31.7$
& \third{$237.5{\pm}9.1$}
& \third{$863.3{\pm}29.5$}
& \second{$631.9{\pm}45.4$} \\

& RIDE
& \second{$717.9{\pm}67.5$}
& \third{$364.9{\pm}117.7$}
& \first{$967.1{\pm}3.3$}
& \second{$239.0{\pm}11.3$}
& \second{$871.3{\pm}22.1$}
& \third{$624.7{\pm}44.0$} \\

& E3B
& \third{$679.2{\pm}199.4$}
& \second{$401.6{\pm}92.1$}
& \third{$961.3{\pm}8.1$}
& $235.4{\pm}24.7$
& $858.0{\pm}22.1$
& $619.5{\pm}40.2$ \\

& HTC$^2$ (ours)
& \first{$827.0{\pm}115.2$}
& \first{$415.8{\pm}111.4$}
& \second{$966.5{\pm}3.8$}
& \first{$248.8{\pm}6.1$}
& \first{$883.8{\pm}12.5$}
& \first{$656.5{\pm}45.0$} \\

\midrule

\multirow{5}{*}{PPO}
& ICM
& $601.9{\pm}179.6$
& \third{$269.1{\pm}47.1$}
& \third{$921.0{\pm}25.4$}
& \second{$89.5{\pm}61.2$}
& \second{$637.6{\pm}55.6$}
& \second{$451.5{\pm}69.2$} \\

& RIDE
& $641.1{\pm}153.5$
& $254.2{\pm}23.1$
& $877.4{\pm}82.0$
& \third{$83.4{\pm}68.5$}
& $565.8{\pm}174.0$
& $425.8{\pm}43.8$ \\

& E3B
& \second{$698.5{\pm}129.7$}
& \second{$277.5{\pm}16.4$}
& $865.7{\pm}100.1$
& $55.3{\pm}65.4$
& \third{$628.1{\pm}53.9$}
& \third{$445.9{\pm}33.3$} \\

& Plan2Explore
& \third{$647.8{\pm}47.3$}
& $239.2{\pm}38.4$
& \second{$929.2{\pm}6.2$}
& $74.2{\pm}68.6$
& $620.7{\pm}51.8$
& $435.2{\pm}37.6$ \\

& HTC$^2$ (ours)
& \first{$733.6{\pm}160.7$}
& \first{$307.9{\pm}22.8$}
& \first{$930.6{\pm}20.1$}
& \first{$100.7{\pm}58.8$}
& \first{$670.2{\pm}25.1$}
& \first{$458.0{\pm}27.4$} \\

\bottomrule
\end{tabular}
}
\end{table}

With SAC, HTC$^2$ achieves the highest mean final return on five of the six tasks. On \texttt{walker\_stand}, HTC$^2$ obtains $966.5\pm3.8$, compared with $967.1\pm3.3$ for RIDE. The largest absolute difference relative to the strongest competing baseline occurs on \texttt{walker\_walk}, where HTC$^2$ reaches $827.0\pm115.2$, compared with $717.9\pm67.5$ for RIDE.

Under PPO, HTC$^2$ achieves the highest mean final return on all six tasks among the evaluated methods, including Plan2Explore. On \texttt{walker\_stand}, HTC$^2$ reaches $930.6\pm20.1$, while Plan2Explore obtains $929.2\pm6.2$, showing very similar mean performance in this high-return regime. On \texttt{hopper\_hop}, HTC$^2$ obtains $100.7\pm58.8$, compared with $89.5\pm61.2$ for the strongest competing baseline, ICM.

Overall, the SAC and PPO results show that the strong mean final-return performance of HTC$^2$ spans multiple control tasks rather than being driven by a single environment. The complete learning curves further show how these differences evolve over training.

%========================================================================================
\subsection{Detailed Physical-State Coverage Results}
\label{app:coverage_details}

We provide the detailed physical-state discretization and numerical statistics corresponding to Figure~\ref{fig:coverage}. All statistics are computed over four random seeds and reported as mean $\pm$ standard deviation.

\paragraph{Coverage-state construction.}
For \texttt{cartpole\_swingup}, we use
\begin{equation}
    s_t^{\mathrm{phy}}
    =
    \left[
        x_{\mathrm{cart},t},
        \operatorname{wrap}
        (\theta_{\mathrm{pole},t})
    \right],
\end{equation}
with a $60\times72$ grid over $x_{\mathrm{cart}}\in[-0.50,0.50]$ and $\theta_{\mathrm{pole}}\in[-\pi,\pi]$.

For \texttt{ball\_in\_cup\_catch}, the physical state is
\begin{equation}
    s_t^{\mathrm{phy}}
    =
    \left[
        x_{\mathrm{ball},t}-x_{\mathrm{cup},t},
        z_{\mathrm{ball},t}-z_{\mathrm{cup},t}
    \right],
\end{equation}
which is discretized using a $64\times64$ grid, with both dimensions restricted to $[-0.35,0.35]$.

For \texttt{finger\_spin}, we use
\begin{equation}
    s_t^{\mathrm{phy}}
    =
    \left[
        \operatorname{wrap}
        (\theta_{\mathrm{obj},t}),
        \operatorname{clip}
        (\dot{\theta}_{\mathrm{obj},t},-25,25)
    \right],
\end{equation}
with a $72\times80$ grid over $\theta_{\mathrm{obj}}\in[-\pi,\pi]$ and $\dot{\theta}_{\mathrm{obj}}\in[-25,25]$. These task-specific physical states are used only for evaluation and do not affect policy learning.

\begin{table*}[t]
\centering
\caption{ \textbf{Physical-state coverage and downstream control performance on three DMC environments.} Final performance and area under the training curve (AUC) are reported as mean $\pm$ standard deviation over four random seeds. For each row, the best and second-best mean values are highlighted in green and yellow, respectively. The highlighting reflects ranking by the reported means. Higher values are better for all metrics. }
\label{tab:coverage_reward_detailed}

\resizebox{\textwidth}{!}{%
\begin{tabular}{llccccc}
\toprule
\textbf{Environment}
& \textbf{Metric}
& \textbf{Statistic}
& \textbf{HTC$^2$ (Ours) + SAC}
& \textbf{ICM + SAC}
& \textbf{RIDE + SAC}
& \textbf{E3B + SAC}
\\
\midrule

\multirow{4}{*}{Cartpole Swingup}
& \multirow{2}{*}{Visited States}
& Final
& \best{$4320.000 \pm 0.000$}
& \second{$3862.250 \pm 486.344$}
& $3759.500 \pm 249.116$
& $3384.750 \pm 399.951$
\\

&
& AUC
& \best{$3525.652 \pm 71.624$}
& $3187.527 \pm 163.460$
& \second{$3200.982 \pm 70.494$}
& $2976.313 \pm 208.753$
\\
\cmidrule(lr){2-7}

& \multirow{2}{*}{Episode Reward}
& Final
& \best{$579.550 \pm 273.080$}
& \second{$126.855 \pm 253.710$}
& $0.000 \pm 0.000$
& $0.000 \pm 0.000$
\\

&
& AUC
& \best{$134.662 \pm 87.884$}
& \second{$15.452 \pm 30.791$}
& $0.058 \pm 0.042$
& $0.017 \pm 0.013$
\\

\midrule

\multirow{4}{*}{Ball in Cup Catch}
& \multirow{2}{*}{Visited States}
& Final
& $2655.000 \pm 55.154$
& \second{$2680.500 \pm 3.536$}
& \best{$2731.500 \pm 14.849$}
& $2664.500 \pm 12.021$
\\

&
& AUC
& \second{$2415.110 \pm 30.751$}
& $2406.878 \pm 69.883$
& \best{$2499.878 \pm 14.418$}
& $2334.732 \pm 19.627$
\\
\cmidrule(lr){2-7}

& \multirow{2}{*}{Episode Reward}
& Final
& \second{$958.730 \pm 2.447$}
& $939.715 \pm 29.013$
& \best{$964.390 \pm 1.909$}
& $945.705 \pm 13.513$
\\

&
& AUC
& \second{$708.565 \pm 87.088$}
& $544.440 \pm 262.849$
& \best{$716.214 \pm 27.763$}
& $399.726 \pm 85.061$
\\

\midrule

\multirow{4}{*}{Finger Spin}
& \multirow{2}{*}{Visited States}
& Final
& \best{$5541.500 \pm 34.648$}
& $5428.500 \pm 58.690$
& $5412.000 \pm 156.978$
& \second{$5513.500 \pm 41.719$}
\\

&
& AUC
& \second{$5317.105 \pm 4.489$}
& $5202.047 \pm 88.931$
& $5195.651 \pm 213.415$
& \best{$5317.360 \pm 34.418$}
\\
\cmidrule(lr){2-7}

& \multirow{2}{*}{Episode Reward}
& Final
& \best{$965.350 \pm 6.604$}
& \second{$957.740 \pm 0.905$}
& $956.460 \pm 3.281$
& $927.550 \pm 41.111$
\\

&
& AUC
& $877.415 \pm 11.824$
& \second{$880.650 \pm 12.565$}
& \best{$881.395 \pm 3.270$}
& $846.177 \pm 28.606$
\\

\bottomrule
\end{tabular}%
}
\end{table*}

Table~\ref{tab:coverage_reward_detailed} reports the complete coverage and reward statistics for the three environments. On \texttt{cartpole\_swingup}, HTC$^2$ reaches $4320.0\pm0.0$ visited states at the end of training and obtains a coverage AUC of $3525.7\pm71.6$, both the highest mean values among the four methods. Its final episode reward and reward AUC are $579.6\pm273.1$ and $134.7\pm87.9$, respectively, which are also the highest reported means for this task.

On \texttt{ball\_in\_cup\_catch}, RIDE achieves the highest final coverage, $2731.5\pm14.8$, followed by ICM at $2680.5\pm3.5$, E3B at $2664.5\pm12.0$, and HTC$^2$ at $2655.0\pm55.2$. In terms of coverage AUC, HTC$^2$ obtains the second-highest mean value of $2415.1\pm30.8$, behind RIDE at $2499.9\pm14.4$. For downstream control, HTC$^2$ reaches a final reward of $958.7\pm2.4$ and a reward AUC of $708.6\pm87.1$, both ranking second among the evaluated methods and remaining close to the corresponding RIDE values of $964.4\pm1.9$ and $716.2\pm27.8$.

On \texttt{finger\_spin}, HTC$^2$ obtains the highest final coverage, $5541.5\pm34.6$, followed by E3B at $5513.5\pm41.7$. The coverage AUC values of HTC$^2$ and E3B are $5317.1\pm4.5$ and $5317.4\pm34.4$, respectively. HTC$^2$ also achieves the highest final reward, $965.4\pm6.6$. For reward AUC, RIDE and ICM obtain $881.4\pm3.3$ and $880.7\pm12.6$, while HTC$^2$ reaches $877.4\pm11.8$.

Across the three tasks, HTC$^2$ achieves the highest final visited-state count on \texttt{cartpole\_swingup} and \texttt{finger\_spin}, while RIDE achieves the highest final coverage on \texttt{ball\_in\_cup\_catch}. The corresponding reward statistics show the same diversity across environments, with HTC$^2$ attaining the highest final reward on \texttt{cartpole\_swingup} and \texttt{finger\_spin}, and the second-highest final reward on \texttt{ball\_in\_cup\_catch}.

\subsection{Additional Hamiltonian Latent Diagnostics}
\label{app:latent_geometry}

This section provides additional diagnostics for the learned Hamiltonian phase representation. The main text shows that the learned scalar $H_\varphi(q,p)$ is stable under passive evolution, responds to policy actions, and varies smoothly over the projected $q$ and $p$ subspaces. Here, we further inspect the internal organization of the phase latents from three complementary perspectives: temporal phase trajectories, action-conditioned phase updates, and dimension-wise $q$--$p$ coupling.

\begin{figure}[h]
% \begin{figure}[ht]
    \centering
    \includegraphics[width=0.82\linewidth]
    {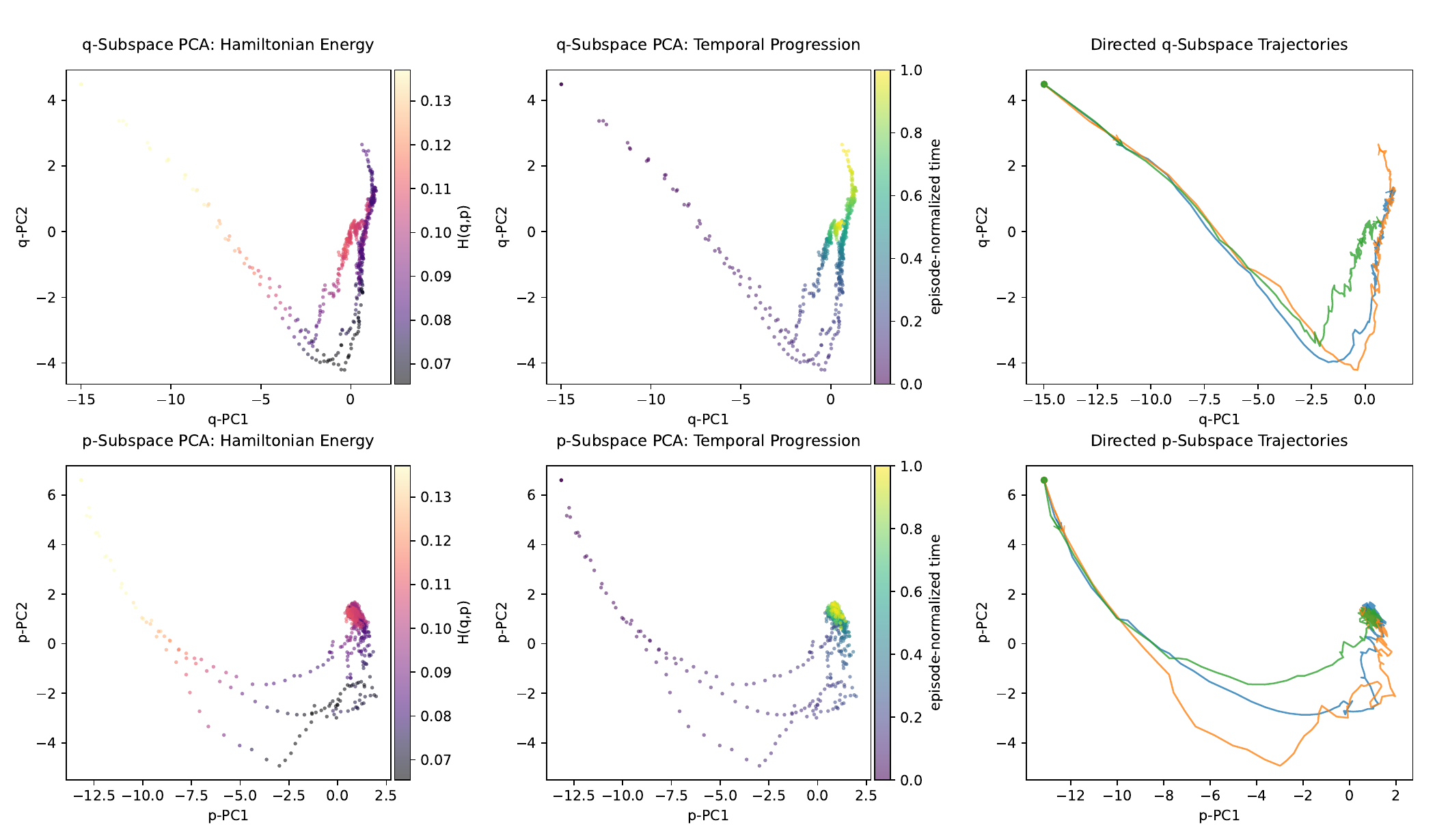}
    \caption{ \textbf{Separate PCA projections of coordinate-like $q$ and momentum-like $p$ latents on \texttt{walker\_walk}.} Each subspace is visualized by learned Hamiltonian energy $H_\varphi(q,p)$, normalized episode time, and projected trajectory traces. The learned phase representation exhibits smooth energy organization and temporally coherent evolution in both subspaces. }
    \label{fig:qp_pca_appendix}
\end{figure}

\paragraph{Phase trajectory organization.}
Figure~\ref{fig:qp_pca_appendix} visualizes the coordinate-like latent $q$ and the momentum-like latent $p$ using separate PCA bases. For each subspace, we show three views: coloring by the learned Hamiltonian energy, coloring by normalized episode time, and projected trajectory traces. Both $q$ and $p$ exhibit smooth energy variation along the projected trajectory, while the normalized-time coloring shows an ordered temporal progression. The trajectory traces further indicate that the learned phase state evolves along coherent low-dimensional paths. This supports the use of posterior phase trajectories for same-episode future pair sampling in HTC$^2$.

\begin{figure}[h]
    \centering
    \includegraphics[width=0.82\linewidth]
    {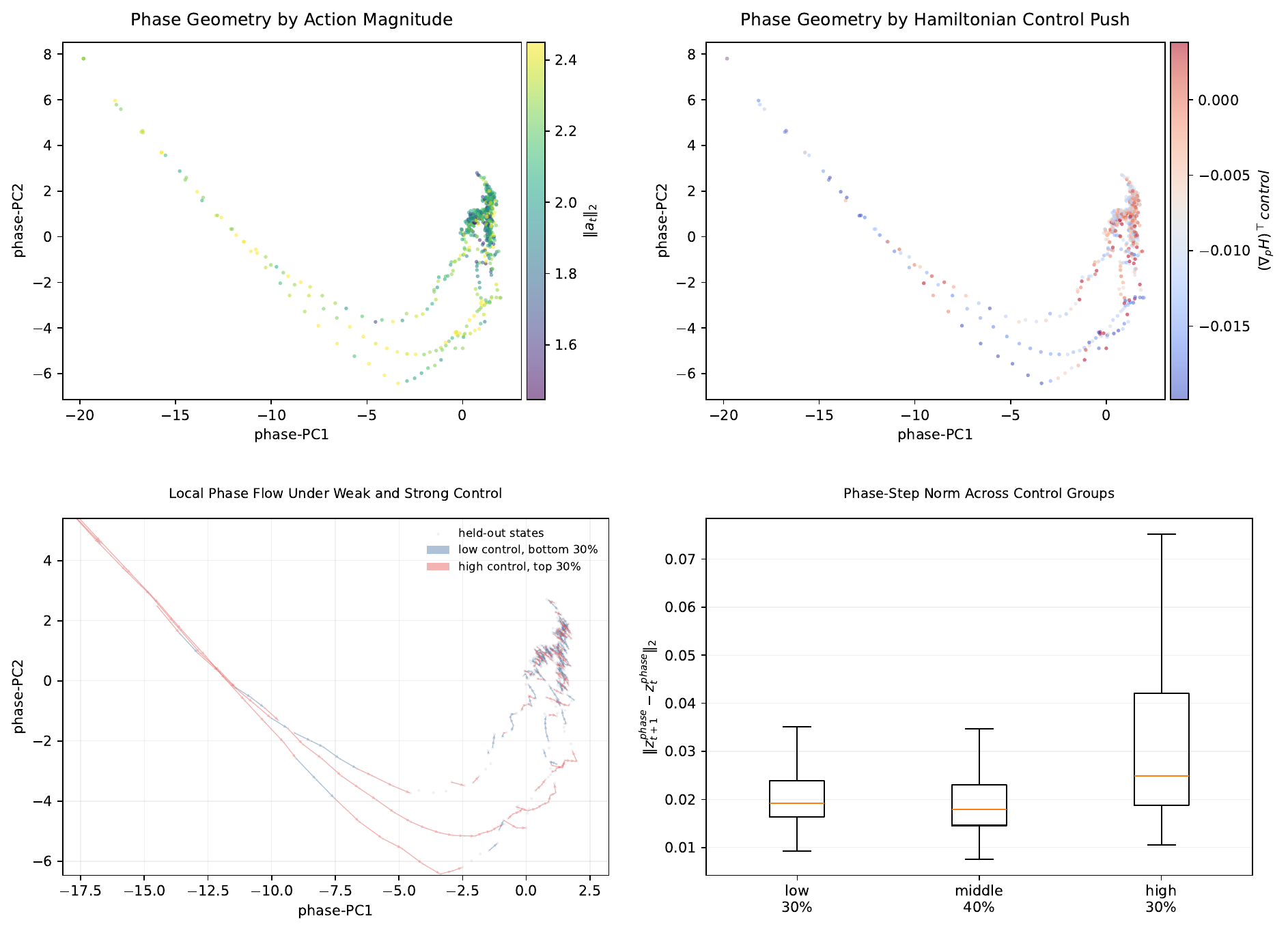}
    \caption{ \textbf{Action-conditioned phase geometry on \texttt{walker\_walk}.} The phase PCA is colored by action magnitude and learned control push. One-step phase updates are grouped by control strength, and the high-control group produces larger phase-step magnitudes. This shows that the learned phase dynamics are modulated by action-conditioned control. }
    \label{fig:action_conditioned_appendix}
\end{figure}

\paragraph{Action-conditioned phase evolution.}
Figure~\ref{fig:action_conditioned_appendix} analyzes how actions modulate the learned phase dynamics. The phase PCA colored by action magnitude and learned control push shows that control-related quantities vary systematically over the phase trajectory rather than being uniformly distributed. The one-step phase updates further separate low-control and high-control transitions: high-control samples produce larger and more directionally distinct phase updates. The phase-step magnitude statistics also increase for the high-control group. These results indicate that the learned phase state is responsive to action-conditioned control, which is the transition structure used by HTC$^2$ to construct horizon-conditioned future pairs.

\begin{figure}[h]
    \centering
    \includegraphics[width=0.82\linewidth]
    % {figure/probe/cross_task_qp_coupling.pdf}
    {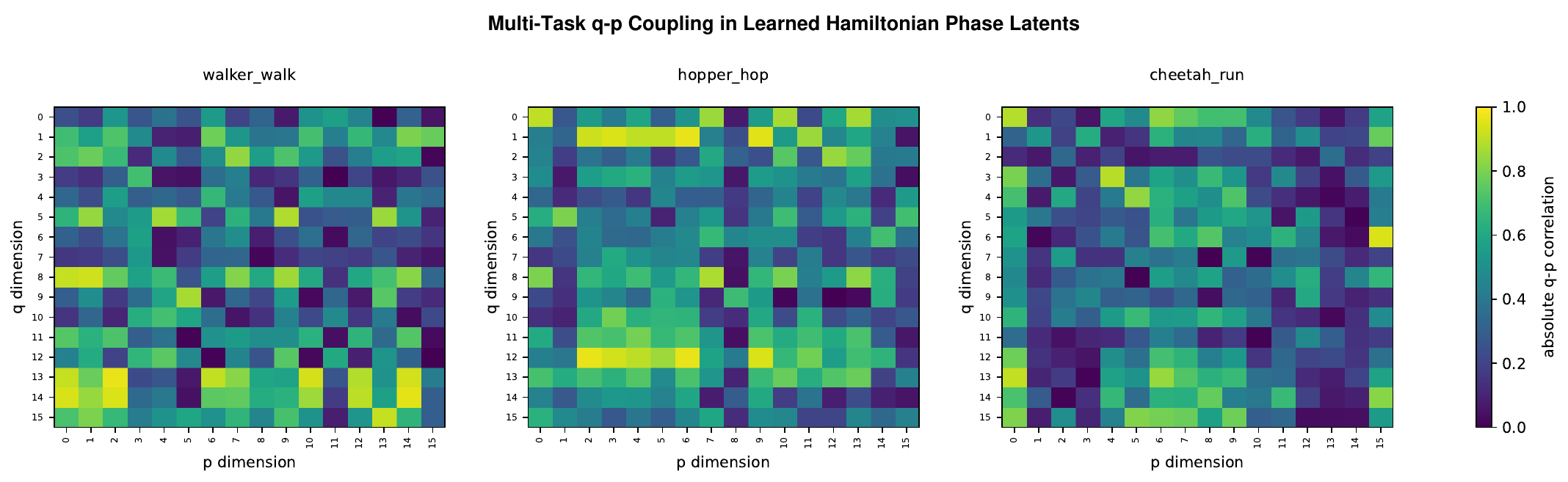}
    \caption{ \textbf{Dimension-wise $q$--$p$ coupling across DMC tasks.} Each heatmap shows the absolute correlation between the 16 dimensions of coordinate-like $q$ and the 16 dimensions of momentum-like $p$. The correlations are local and dimension-selective rather than uniformly high, indicating that $q$ and $p$ remain dynamically related but do not collapse into a single fully mixed latent representation. }
    \label{fig:qp_coupling_appendix}
\end{figure}

\paragraph{Dimension-wise $q$--$p$ coupling.}
Figure~\ref{fig:qp_coupling_appendix} reports the dimension-wise absolute correlation between the coordinate-like subspace $q$ and the momentum-like subspace $p$ on \texttt{walker\_walk}, \texttt{hopper\_hop}, and \texttt{cheetah\_run}. Instead of treating $q$ and $p$ as two monolithic vectors, this analysis inspects all $16\times16$ cross-dimensional pairs. The heatmaps show that the $q$--$p$ relation is not uniformly coupled across all dimensions: strong correlations appear only for selected dimension pairs, while many other pairs remain weakly or moderately correlated. This pattern indicates that the learned phase representation does not collapse $q$ and $p$ into a single mixed latent block. Rather, the two subspaces share the same trajectory-level dynamics while preserving a degree of dimension-wise separation.

This result is consistent with the intended phase-space structure of HTC$^2$. The coordinate-like and momentum-like components should not be fully independent, because they are inferred from the same controlled trajectory and jointly determine the learned Hamiltonian $H_\varphi(q,p)$. At the same time, they should not become identical or uniformly entangled. The observed local and dimension-selective coupling therefore supports the use of a structured phase representation for Hamiltonian temporal contrastive exploration.

\paragraph{Summary.}
These diagnostics provide a finer view of the Hamiltonian phase representation used by HTC$^2$. The separate $q/p$ PCA plots show energy-ordered and time-coherent phase trajectories; the action-conditioned analysis shows that phase updates are modulated by control strength; and the cross-task $q$--$p$ coupling maps show trajectory-level dependence without uniform dimension-wise collapse. Together, these results support the representational structure assumed by HTC$^2$: exploration is driven by temporal future discovery in a Hamiltonian phase space whose components are dynamically related yet not reduced to a single entangled latent cloud.
%============================================================================================================================
\subsection{Detailed Parametric OOD Results}
\label{app:ood_details}

This section reports the detailed numerical values for the parametric OOD policy-retention experiment in Section~\ref{subsec:ood}. Retention is computed as
\[
    \mathrm{Retention}
    =
    100 \times
    \frac{R_{\mathrm{OOD}}}{R_{\mathrm{ID}}},
\]
where $R_{\mathrm{OOD}}$ is the zero-shot return under the perturbed environment and $R_{\mathrm{ID}}$ is the return in the original environment for the same method and seed. We evaluate four physical shifts: gravity ($1.5\times$), friction ($2.0\times$), mass ($1.5\times$), and damping ($2.0\times$).

\begin{table}[h]
\centering
\caption{ Parametric OOD zero-shot policy retention on \texttt{walker\_walk}. Values report return retention relative to the base environment (\%, mean $\pm$ std). }
\label{tab:ood_walker_walk}
\small
\begin{tabular}{lcccc}
\toprule
Model
& Gravity 1.5$\times$
& Friction 2.0$\times$
& Mass 1.5$\times$
& Damping 2.0$\times$ \\
\midrule
SAC+ICM
& 69.58 $\pm$ 3.02
& 94.10 $\pm$ 6.05
& 64.64 $\pm$ 13.19
& 101.45 $\pm$ 0.08 \\
SAC+RIDE
& 58.79 $\pm$ 8.01
& 95.85 $\pm$ 0.21
& 62.80 $\pm$ 13.66
& \textbf{101.73 $\pm$ 0.13} \\
SAC+E3B
& 70.08 $\pm$ 16.26
& 97.36 $\pm$ 1.56
& 70.65 $\pm$ 14.16
& 100.46 $\pm$ 1.79 \\
SAC+HTC$^2$
& \textbf{78.70 $\pm$ 11.94}
& \textbf{100.01 $\pm$ 0.83}
& \textbf{82.38 $\pm$ 10.40}
& 97.68 $\pm$ 3.05 \\
\bottomrule
\end{tabular}
\end{table}

\begin{table}[h]
\centering
\caption{ Parametric OOD zero-shot policy retention on \texttt{hopper\_stand}. Values report return retention relative to the base environment (\%, mean $\pm$ std). }
\label{tab:ood_hopper_stand}
\small
\begin{tabular}{lcccc}
\toprule
Model
& Gravity 1.5$\times$
& Friction 2.0$\times$
& Mass 1.5$\times$
& Damping 2.0$\times$ \\
\midrule
SAC+ICM
& 5.42 $\pm$ 6.95
& 99.11 $\pm$ 1.21
& 40.76 $\pm$ 0.90
& 99.45 $\pm$ 1.37 \\
SAC+RIDE
& \textbf{52.37 $\pm$ 14.84}
& 96.55 $\pm$ 1.28
& 64.32 $\pm$ 15.83
& \textbf{100.44 $\pm$ 0.30} \\
SAC+E3B
& 4.23 $\pm$ 3.48
& 94.85 $\pm$ 2.71
& 32.28 $\pm$ 17.39
& 99.60 $\pm$ 0.23 \\
SAC+HTC$^2$
& 30.21 $\pm$ 5.63
& \textbf{99.15 $\pm$ 1.36}
& \textbf{85.73 $\pm$ 10.99}
& 97.65 $\pm$ 3.06 \\
\bottomrule
\end{tabular}
\end{table}

\begin{table}[h]
\centering
\caption{ Parametric OOD zero-shot policy retention on \texttt{cheetah\_run}. Values report return retention relative to the base environment (\%, mean $\pm$ std). }
\label{tab:ood_cheetah_run}
\small
\begin{tabular}{lcccc}
\toprule
Model
& Gravity 1.5$\times$
& Friction 2.0$\times$
& Mass 1.5$\times$
& Damping 2.0$\times$ \\
\midrule
SAC+ICM
& 70.93 $\pm$ 2.46
& 54.84 $\pm$ 25.55
& 89.33 $\pm$ 17.98
& 43.74 $\pm$ 9.29 \\
SAC+RIDE
& 77.72 $\pm$ 9.16
& \textbf{92.31 $\pm$ 8.88}
& 100.90 $\pm$ 7.78
& 63.70 $\pm$ 4.68 \\
SAC+E3B
& 74.47 $\pm$ 27.46
& 61.43 $\pm$ 7.62
& 96.43 $\pm$ 18.18
& \textbf{67.63 $\pm$ 6.52} \\
SAC+HTC$^2$
& \textbf{84.18 $\pm$ 2.15}
& 64.67 $\pm$ 38.44
& \textbf{107.85 $\pm$ 8.09}
& 59.89 $\pm$ 4.71 \\
\bottomrule
\end{tabular}
\end{table}

\paragraph{Discussion.}
The detailed results show that HTC$^2$ obtains the best retention on seven out of twelve OOD settings. The strongest pattern appears under mass perturbations, where HTC$^2$ achieves the highest retention on all three tasks. It also achieves the best gravity-shift retention on \texttt{walker\_walk} and \texttt{cheetah\_run}. These results are consistent with the objective of HTC$^2$: exploration is guided by multi-step temporal discovery in a Hamiltonian phase representation, which encourages policies to experience physically informative transitions during training. Friction and damping shifts show more task-dependent behavior, indicating that different intrinsic rewards can favor different robustness profiles.

% ========================================================================================================%
\subsection{Ablation Learning Curves}
\label{app:ablation_learning_curves}

\begin{figure}[t]
    \centering
    \includegraphics[width=\linewidth]{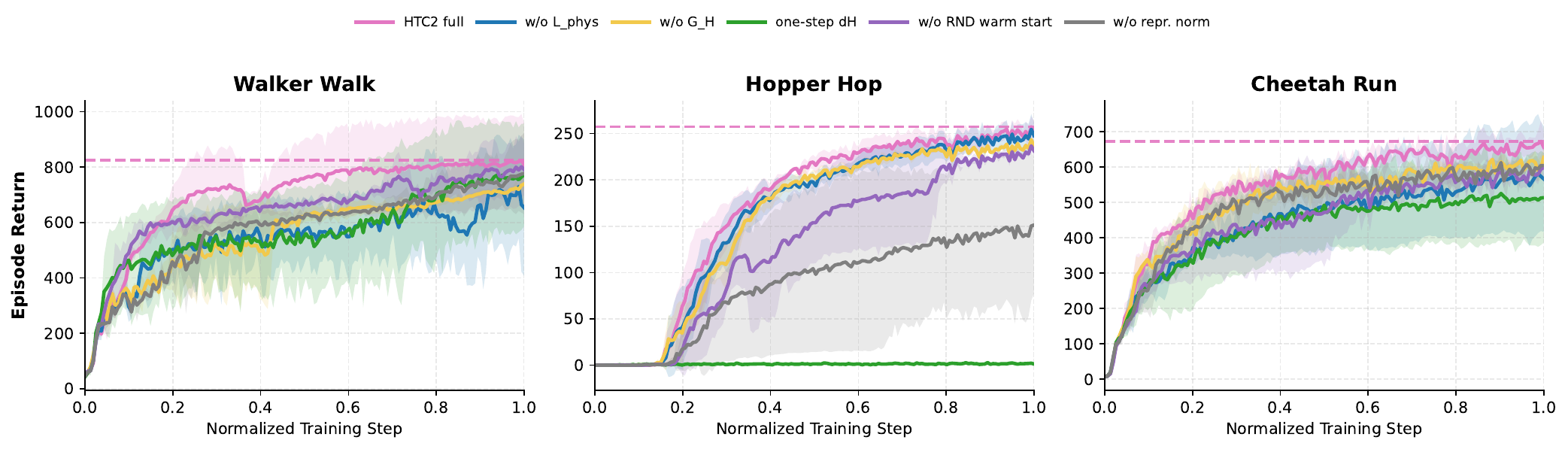}
    \caption{ \textbf{Ablation learning curves.} Curves report mean performance over 4 seeds, and shaded regions denote standard deviation. Full HTC$^2$ is compared with variants removing the physical consistency loss, Hamiltonian energy gate, RND warm start, representation normalization, and with a one-step $\Delta H$ reward. }
    \label{fig:htc2_ablation_curves}
\end{figure}

We further analyze the sample efficiency of the ablation variants in Figure~\ref{fig:htc2_ablation_curves}. The curves compare full HTC$^2$ with variants that remove the physical consistency loss, remove the Hamiltonian energy gate, replace temporal contrastive reward with one-step $\Delta H$, remove the RND~\citep{burda2018exploration} warm start, or remove representation normalization.

On \texttt{walker\_walk}, full HTC$^2$ improves steadily and reaches the highest final return. Several variants learn useful behaviors, but the full method maintains the strongest late-stage performance.

On \texttt{hopper\_hop}, the difference between variants is more pronounced: full HTC$^2$ reaches high return and remains stable, whereas the one-step $\Delta H$ reward fails to learn an effective hopping behavior. The variant without representation normalization also suffers a large drop, indicating that stable contrastive distances are important for this underactuated task.

On \texttt{cheetah\_run}, full HTC$^2$ again achieves the strongest final performance. The variants without $\mathcal{L}_{\mathrm{phys}}$, without the energy gate, or without RND~\citep{burda2018exploration} warm start improve during training but converge to lower returns. These learning curves are consistent with the final-return ablation in Table~\ref{tab:htc2_ablation}: the complete HTC$^2$ design improves both sample efficiency and final performance by combining Hamiltonian phase learning with multi-step temporal contrastive curiosity.

\subsection{Policy Rollout Visualization}
\label{app:policy_rollout_visualization}

\begin{figure*}[t]
    \centering
    \includegraphics[width=\textwidth]
    {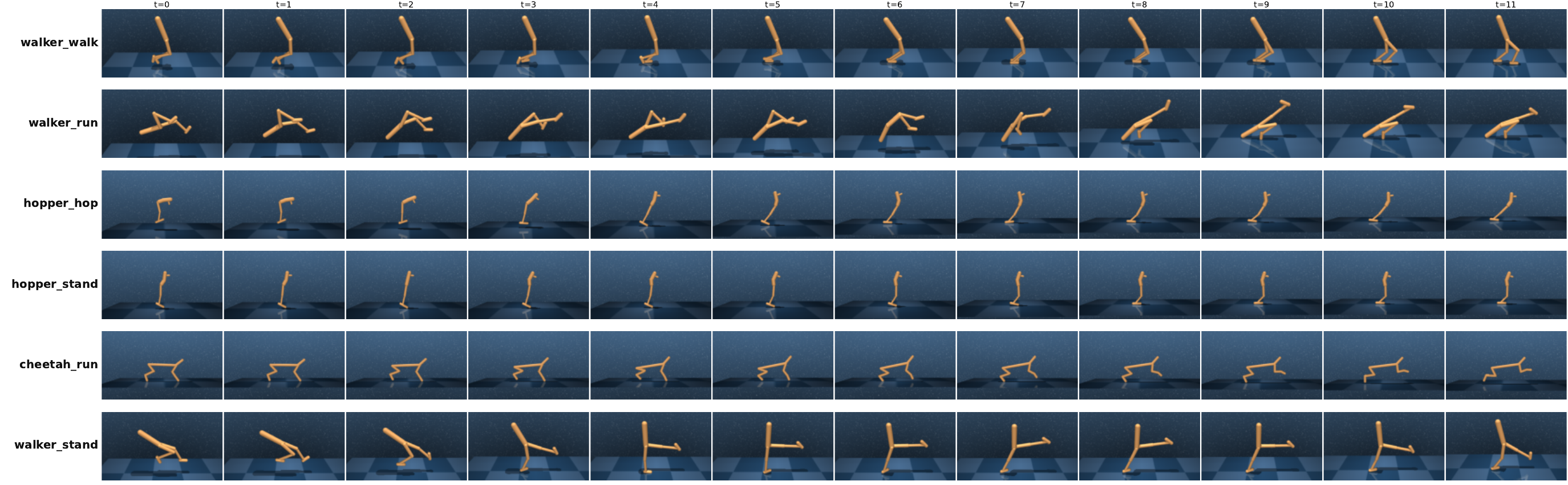}
    \caption{ \textbf{HTC$^2$ policy rollouts on six DMC tasks.} Each row corresponds to one task, and columns show twelve temporally ordered sampled frames ($t=0,\ldots,11$) from the final trained policy checkpoint. The visualization is used only as a qualitative inspection of policy behavior. }
    \label{fig:htc2_rollout_summary}
\end{figure*}

To qualitatively inspect the behavior induced by the final HTC$^2$ policy, we perform additional environment rollouts from the trained checkpoint and sample temporally ordered observations from six DMC tasks. Figure~\ref{fig:htc2_rollout_summary} shows twelve sampled frames ($t=0,\ldots,11$) for \texttt{walker\_walk}, \texttt{walker\_run}, \texttt{hopper\_hop}, \texttt{hopper\_stand}, \texttt{cheetah\_run}, and \texttt{walker\_stand}.

The sampled trajectories exhibit distinct task-dependent motion patterns. On \texttt{walker\_walk}, the agent remains upright while its leg configuration evolves across the rollout, whereas \texttt{walker\_run} exhibits substantially larger changes in body and limb configuration. The hopper remains close to an upright posture on \texttt{hopper\_stand}, while \texttt{hopper\_hop} shows greater temporal variation in its articulated configuration. On \texttt{cheetah\_run}, the torso and limbs evolve through a sequence of locomotion configurations. For \texttt{walker\_stand}, the rollout progresses from an initially inclined configuration toward a more upright posture, followed by continued balancing adjustments.